\documentclass[conference]{IEEEtran}
\IEEEoverridecommandlockouts
\usepackage{cite}
\usepackage{amsmath,amssymb,amsfonts}
\usepackage{graphicx}
\usepackage{textcomp}
\usepackage{xcolor}
\def\BibTeX{{\rm B\kern-.05em{\sc i\kern-.025em b}\kern-.08em
    T\kern-.1667em\lower.7ex\hbox{E}\kern-.125emX}}
\newcommand*\titleheader[1]{\gdef\@titleheader{#1}}
\usepackage{amsthm}%
\usepackage{algorithm}%
\usepackage{algorithmicx}%
\usepackage[noend]{algpseudocode}
\usepackage{hyperref}       % hyperlinks
\usepackage{url}            % simple URL typesetting
\usepackage{booktabs}       % professional-quality tables
\usepackage{amsmath,amsfonts,amssymb}
\usepackage{doi}
\ifCLASSOPTIONcompsoc
    \usepackage[caption=false, font=normalsize, labelfont=sf, textfont=sf]{subfig}
\else
\usepackage[caption=false, font=footnotesize]{subfig}
\fi

\usepackage{tikz}
\algnewcommand{\LineComment}[1]{\State $\triangleright$ #1}
\algnewcommand{\pia}[1]{$\pi_{\text{adv}}$}

\DeclareMathOperator*{\argmin}{arg\,min}

\usepackage{ifthen}
\usepackage{bm}
\usepackage{amsfonts}
\usepackage{stackengine}
\usepackage{cleveref}
\usetikzlibrary{arrows.meta,arrows}
\let\oldhref\href
\renewcommand{\href}[2]{\oldhref{#1}{\hbox{#2}}}

\usepackage[british]{babel}
\usepackage{hyphenat}

\newtheorem{theorem}{Theorem}

\makeatletter 

\let\old@ps@IEEEtitlepagestyle\ps@IEEEtitlepagestyle
\def\confheader#1{%
    \def\ps@IEEEtitlepagestyle{%
        \old@ps@IEEEtitlepagestyle%
        \def\@oddhead{\strut\hfill#1\hfill\strut}%
        \def\@evenhead{\strut\hfill#1\hfill\strut}%
    }%
    \ps@headings%
}
\makeatother

\confheader{%
        \parbox{20cm}{\textbf{\textit{Accepted at IEEE Conference on Artificial Intelligence (CAI 2024).}}}
}

\begin{document}

\title{Robust Lagrangian and Adversarial Policy Gradient for Robust Constrained Markov Decision Processes \thanks{This work has been supported by the UKRI Trustworthy Autonomous Systems Hub, EP/V00784X/1, and was part of the Safety and Desirability Criteria for AI-controlled Aerial Drones on Construction Sites project.}}
\author{\IEEEauthorblockN{1\textsuperscript{st} David M. Bossens}
\IEEEauthorblockA{\textit{IHPC}\\
 \textit{Agency for Science, Technology and Research}\\ 
 Singapore\\
\textit{CFAR}\\ 
\textit{Agency for Science, Technology and Research}\\ 
Singapore \\
bossensdm@cfar.a-star.edu.sg}}
\maketitle

\begin{abstract}
Robustness and safety constraints are key requirements for AI systems. The robust constrained Markov decision process is a recent task-modelling framework that incorporates behavioural constraints and robustness to reinforcement learning systems. Earlier work proposed the robust constrained policy gradient (RCPG) algorithm, which robustifies either the value or the constraint and updates the worst-case distribution through constrained optimisation on a sorted value list. Highlighting potential downsides of RCPG such as not robustifying the full constrained objective and the lack of incremental learning, this paper introduces algorithms to robustify the Lagrangian and to learn incrementally using gradient descent over an adversarial policy. A theoretical analysis derives the Lagrangian policy gradient for the policy optimisation and the Lagrangian adversarial policy gradient for the adversary optimisation. Empirical experiments injecting perturbations in inventory management and safe navigation tasks demonstrate the benefit of these modifications, and combining both modifications yields the best overall performance.
\end{abstract}


\section{Introduction}
Reinforcement learning (RL) is the standard framework for interactively learning in a complex environment. By maximising a long-term utility function, traditional RL does not take into account various behavioural constraints that would be desirable for a policy (e.g. to ensure safety or to follow legal and moral norms). Moreover, RL systems typically assume that the agent can learn directly in the true transition dynamics model. This is often not the case: for instance, in applications such as robotic control and recommendation, one may want to learn from a simulated environment rather than the true environment as this will be more safe.

Due to allowing to learn policies that satisfy long-term behavioural constraints, constrained Markov decision processes (CMDPs) \cite{Altman1998} have become the de facto standard for safe reinforcement learning \cite{Gu2022}. CMDPs formulate a constraint-cost function in addition to the reward function and formalise long-term behavioural constraints based on a threshold of the expected cumulative constraint-cost.

Tied to safety is the concept of robustness, which is the ability to retain performance even when the true environment changes or differs from the training environment. Robustness is represented in RL by robust Markov decision processes (RMDPs) \cite{Iyengar2005,Nilim2005}, which conceptualise robustness in terms of the uncertainty over the transition dynamics model of the MDP. While there are alternatives for robust RL such as domain randomisation \cite{VanBaar2019} and meta-learning \cite{Finn2017}, these solutions are less theoretically sound.

The recently proposed framework of robust CMDPs (RCMDPs) \cite{Russel2021} optimises the CMDP with the worst-case dynamics model in the uncertainty set, effectively combining RMDPs and CMDPs. To optimise RCMDP policies, the Robust Constrained Policy Gradient (RCPG) algorithm \cite{Russel2021} combines a policy gradient algorithm with a Lagrangian relaxation for constraints and a worst-case dynamics computation for robustness. RCPG regularly recomputes the worst-case dynamics by sorting the list of values from each state and then performing constrained minimisation of the value subject to the norm constraints (i.e. the maximal distance to the expected, or ``nominal'', dynamics). The algorithm may not be optimal for learning robust policies since a) the algorithm does not consider the combined objective of rewards and constraint-costs; b) immediately presenting worst-case dynamics may prevent learning important and representative patterns; and c) due to repeatedly performing constrained optimisation over a sorted value list, the transition distributions of all state-action pairs can be subject to large changes whenever a state has a changed value estimate.

To mitigate these three problems, this paper proposes two algorithms. First, mitigating problem a), a variant of RCPG is introduced, called RCPG with Robust Lagrangian, which computes the worst-case over the Lagrangian, which combines the expected cumulative reward with the expected cumulative constraint-cost into a single objective. Second, to mitigate all three problems, an algorithm called Adversarial RCPG is proposed, which uses an adversary to minimise the Lagrangian of the current RL policy subject to the constraints of the uncertainty set (e.g. the L1 distance to the nominal model). Adversarial RCPG thereby addresses the above-mentioned limitations of RCPG by using a worst-case Lagrangian objective and by incrementally updating an adversary that represents the dynamics directly -- rather than updating the dynamics indirectly and abruptly based on a sorted value list.

\section{Preliminaries}
\label{sec: preliminaries}
The RCMDP framework is defined by a tuple $(\mathcal{S},\mathcal{A},r,c,d,\gamma, P_*, \mathcal{P})$, where $\mathcal{S}$ is the state space, $\mathcal{A}$ is the action space, $r$ is the reward function, $c$ is the constraint-cost function, $d$ is the budget of expected cumulative constraint-cost, $\gamma$ is the discount factor, $P_*$ is the unknown true transition dynamics model, and $\mathcal{P}$ is the uncertainty set which includes many candidate transition dynamics models. The value of executing a policy from a given state $s \in \mathcal{S}$ given a particular transition model $P \in \mathcal{P}$ is given by the expected discounted cumulative reward,
\begin{equation}
V_{\pi,P}(s) = \mathbb{E}\left[\sum_{t=0}^{\infty} \gamma^t r(s_t) \vert s_0=s, a_t \sim \pi(s_t), s_{t+1}\sim P_{s_{t},a_{t}}\right] \,.
\end{equation} 
Analogously, the expected discounted cumulative constraint-cost is denoted by
\begin{equation}
C_{\pi,P}(s) = \mathbb{E}\left[\sum_{t=0}^{\infty} \gamma^t c(s_t) \vert s_0=s, a_t \sim \pi(s_t), s_{t+1}\sim P_{s_{t},a_{t}}\right] \,.
\end{equation} 
Denoting $P^{+} := \min_{P \in \mathcal{P}} V_{\pi,P}(s)$, the objective within the RCMDP framework is given by
\begin{equation}
\label{eq: RCMDP objective}
\max_{\pi} V_{\pi,P^{+}}(s) \quad 
\text{s.t.}  \quad C_{\pi,P^{+}}(s) \leq  d \,.
\end{equation}
Note that instead of the worst-case over the value, $P^{+}$ may also be defined as the worst-case over the constraint-cost.

\section{Related work}
\label{sec: related work} 
RCMDPs are a recent field of endeavour with few directly related works. Below section summarises the directly related works as well as works combining CMDPs with other notions of robustness.
\subsection{RCMDP related works}
Russel, Benosman, and Van Baar \cite{Russel2021,Russel2022} formulate RCMDPs as defined in Sec.~\ref{sec: preliminaries}, effectively combining distributional robustness with constrained Markov decision processes. They propose the Robust Constrained Policy Gradient (RCPG), which is a policy gradient algorithm that uses L1-norm uncertainty sets and Lagrange relaxation \cite{Bertsekas2003}. While Lagrange relaxation is common in CMDP works (e.g. \cite{Achiam2017,Ray2019}), the algorithm additionally provides worst-case distributional robustness over the L1-norm uncertainty set. The worst-case dynamics distribution for a given state-action pair is computed by first formulating a list of sorted values or constraint-costs and then applying linear programming or a special-purpose algorithm to re-assign probabilities subject to the norm constraints \cite{Petrik2005}. Lyapunov-based reward shaping has also been shown to yield convergence to a local optimum, although its benefits have not been demonstrated in practice \cite{Russel2022}. Adversarial RCPG modifies the RCPG algorithm by replacing the worst-case value computation with an adversarial training scheme. In this scheme, the adversary provides transition dynamics that minimise the Lagrangian of the policy's RCMDP objective. The training is incremental,  starting from the nominal model and gradually changing to less representative and more difficult CMDPs. These features help provide learning progress as well as robustness to the full constrained objective.

Other works related to RCMDPs are tailored to somewhat different purposes.  Explicit Explore, Exploit, or Escape (E4) \cite{Bossens2022} provides a framework for safe exploration in RCMDPs. The approach distinguishes between known states, where the CMDP model and therefore the value function is approximately correct, and unknown states, where a worst-case assumption is taken on the transitions and the constraint-cost. The approach yields near-optimal policies for the underlying CMDP in polynomial time while satisfying the constraint-cost budget at all times. While the approach has solid theoretical support, maintaining safety throughout exploration is not always the primary concern and comes at significant training costs. The present paper is mainly interested in the final policy being safe rather than in safe exploration, and assumes an uncertainty set is available at the start of learning. The R3C objective  \cite{Mankowitz2020a} combines the worst-case value and the worst-case constraint over distinct simulators with different parametrisations being run for one step from the current state. Avoiding to explicitly compute the transition dynamics matrix makes it applicable to large scale domains such as control problems. However, the number of simulators must be limited (e.g. 4 distinct transition dynamics), thereby reducing the worst-case robustness and the scope of robustness. The present paper focuses on L1 uncertainty sets derived from state-action trajectories; such sets include a much wider range of dynamics and do not require significant prior knowledge of the environment (e.g. the internal parameters of a simulator).

\subsection{Other approaches to robustifying CMDPs}
Other approaches propose techniques other than worst-case optimisation to robustify CMDPs. Some works assume that transition dynamics are known but the reward and constraint functions are not. For instance, Zheng et al. (2020) \cite{Zheng2020} have previously used a robust version of LP in the context of UCRL, which estimates an upper confidence bound on the cost and the reward. Another approach is to use the conditional value at risk (CVaR); for instance, PG-CVaR and AC-CVaR \cite{Chow2014} consider a CVaR of the value function (in a non-constrained approach) and later approaches use the CVaR for defining a cost critic for the CMDP \cite{Yang2021,Zhang2023}. Other works have also explored the use of Lyapunov stability to ensure safe exploration within CMDPs \cite{Chow2019}. Concepts of stability and safe exploration are complementary to RCMDPs, and indeed have been investigated in theory but not in practice \cite{Russel2022,Bossens2022}. Compared to these exemplary approaches, the RCMDP framework focuses on the uncertainty in dynamics models, making it particularly useful when transition dynamics are estimated from observational data. Other approaches define robustness guarantees based on an available baseline policy. For instance, SPIBB considers safe policy improvement across the uncertainty set in the sense of guaranteeing at least the performance of a baseline policy \cite{Laroche2019}; as shown in Satija et al. \cite{Satija2021}, this approach can be cast in a CMDP framework.

\section{Adversarial RCPG}
\label{sec: AdvRCPG}
Adversarial RCPG modifies RCPG by using a function approximator for the worst-case dynamics and by combining the values and constraints into a single objective. This is achieved by updating, by policy gradient, an adversary $\pi_{\text{adv}}$ as the dynamics that minimise the Lagrangian that the policy $\pi$ is maximising. Before explaining the details of Adversarial RCPG, this section first presents the original RCPG algorithm. 
\subsection{Robust-Constrained Policy Gradient}
\label{sec: RCPG}
RCPG finds the saddle point of the Lagrangian for a given budget $d$. Denoting $\pi_{\theta}$ as the policy parametrised by $\theta$, $\lambda$ as the Lagrangian multiplier, and $P^{+}$ as the worst-case transition dynamics, the objective is given by
\begin{equation}
\label{eq: Lagrange}
\min_{\lambda \geq 0} \max_{\pi_{\theta}} L(\lambda,\pi_{\theta}; P^{+}) =  V_{\pi_{\theta},P^{+}}(s) - \lambda \left(C_{\pi_{\theta},P^{+}}(s)    - d\right) \,.
\end{equation}
To optimise the above objective, sampling of limited-step trajectories is repeated for a large number of independent iterations starting from a random initial state $s \sim P_0$. Based on the large number of trajectories collected, one then performs gradient descent in $\lambda$ and gradient ascent in $\theta$. 

\paragraph{Estimating the worst-case distribution}
RCPG is based on L1 uncertainty sets of the type $\mathcal{P}_{s,a}=\{ P \in \Delta^S : || P - \hat{P}_{s,a} ||_1 \leq \alpha \}$. Computing the worst-case distribution, also known as the ``inner problem'', is equivalent to the constrained optimisation problem
\begin{align}
\label{eq: robustvalue}
P^{+} = \argmin_P V^+ = P^{\intercal} \hat{V} \quad \text{s.t.} & \quad || P - \hat{P}_{s,a} ||_1 \leq \alpha \\
											  & \quad \mathbf{1}^{\intercal} P = 1 \,. \notag
\end{align}
The solution to the inner problem, $P^{+}$, is the distribution that minimises the value subject to the norm constraints. 
RCPG solves the innner problem based on linear programming or related constrained optimisation algorithms (e.g. Petrik et al. \cite{Petrik2005} present a special purpose algorithm that solves the problem with $O(S \log S)$ time complexity). These algorithms require a tabular approximation or (preferably) a critic network of the quantity to robustify (the expected cumulative reward or constraint-cost).

\paragraph{Learning problems of RCPG}
The RCPG algorithm has several features in its learning that could be improved. First, the RCPG objective provides robustness to either the worst-case value or the worst-case constraint-cost but not the desired Lagrangian objective combining both. Second, RCPG training results in only training on the worst case, which may be too challenging at the start of training; a more gradual training with an incrementally improving adversary could be beneficial. Third, if the critic estimates the worst-case state erroneously then the updated worst-case transition dynamics will sample this state at an excessively high rate as the next state for all the state-action pairs. This results in many abrupt changes in the distributions, again making an incremental learning process difficult.
\subsection{RCPG with Robust Lagrangian}
\label{sec: Lagrangian RCPG}
A relatively straightforward way to solve the first learning problem of RCPG is to directly robustify the Lagrangian Eq.~\ref{eq: Lagrange}. RCPG with Robust Lagrangian replaces Eq.~\ref{eq: robustvalue} with 
\begin{align}
\label{eq: robustlagrangian}
P^{+} = \argmin_P L^+ = P^{\intercal} \left( \hat{V} - \lambda \hat{C}\right) \quad \text{s.t.} & \quad || P - \hat{P}_{s,a} ||_1 \leq \alpha \\
											  & \quad \mathbf{1}^{\intercal} P = 1 \,. \notag
\end{align}
As shown in Theorem~\ref{th: policy grad}, the same RCPG algorithm can be used for optimising Eq.~\ref{eq: Lagrange} when the worst-case distribution $P^{+}$ is defined based on Eq.~\ref{eq: robustlagrangian}.
\subsection{Adversarial RCPG}
While RCPG with Robust Lagrangian provides a suitable objective, it does not address the other learning problems of RCPG. The Adversarial RCPG algorithm is proposed to mitigate all three learning problems. It learns an adversarial policy $\pi_{\text{adv}} : \mathcal{S} \times \mathcal{A} \to \Delta^{S}$ which approximates the robust Lagrangian dynamics model $P^+$ from Eq.~\ref{eq: Lagrange} to robustify the full constrained objective. The adversarial policy is learned by gradient descent together with the policy: the adversary starts from the nominal distribution $\hat{P}$ and incrementally updates the dynamics to be more challenging using gradient descent. The resulting algorithm is shown in Algorithm~\ref{alg: Adversarial RCPG}).

As in Lagrangian RCPG, Adversarial RCPG optimises the Lagrangian in Eq.~\ref{eq: Lagrange} robustly based on the worst-case distribution defined in Eq.~\ref{eq: robustlagrangian}. However, Adversarial RCPG replaces $P^+$ with an adversarial policy $\pi_{\text{adv}}$ which minimises the policy's objective subject to the norm constraints:
\begin{align}
\pi_{\text{adv}} = &\argmin_{\pi_{\text{adv}}} L(\lambda,\pi_{\theta}; \pi_{\text{adv}}) \\
&\quad \text{s.t.} \quad ||\pi_{\text{adv}}(s,a) - 
\hat{P}_{s,a} ||_1 \leq \alpha(s,a) \forall (s,a) \in \mathcal{S} \times \mathcal{A} \,. \notag
\end{align}
This leads to a constrained optimisation problem where $\pi_{\text{adv}}$ is the solution to a different Lagrangian $L_{\text{adv}}(\lambda_{\text{adv}},\pi_{\text{adv}})$, namely
\begin{align}
\max_{\lambda_{\text{adv}} \geq 0} \min_{\pi_{\text{adv}}} & L_{\text{adv}}(\lambda_{\text{adv}},\pi_{\text{adv}}) = L(\lambda,\pi_{\theta}; 
\pi_{\text{adv}}) \\ & + \sum_{s,a} \lambda_{\text{adv}} \left( ||\pi_{\text{adv}}(s,a) - \hat{P}_{s,a} ||_1 - \alpha(s,a) \right) \,, \notag
\end{align}
where the multiplier is the same for all state-action pairs to make the optimisation scalable to large state-action spaces.

The distribution of states depends crucially on the adversary, and this distribution should be within an L1-norm of $\alpha(s,a)$ for all $(s,a) \in \mathcal{S} \times \mathcal{A}$. Minimising the L1 norm based on samples obtained from $\pi_{\text{adv}}$ itself may lead to a system hack, in the sense that the adversarial policy may learn to sample states which have minimal L1 norm rather than minimising the L1 norm independent of the state-action pairs observed. Therefore, to compute the gradient for the L1 norm (l.20) for the adversarial policy update, a new random batch with random state-action samples is used for each gradient such that overall the norm is being reduced regardless of the state-action pairs encountered (see Algorithm~\ref{alg: deviation-from-nominal}). By contrast, the term $\Delta P$ (l.22) is used for updating the multiplier only, so there is no risk for such system hacks; therefore, it is based directly based on the samples from the adversary, which gives more impact of frequently observed state-action pairs on the multiplier. This approach empirically satisfies the norm constraints.

\begin{algorithm}
\caption{Offline optimisation with Adversarial RCPG} \label{alg: Adversarial RCPG}
\begin{minipage}[c]{0.49\textwidth}
\small
\begin{algorithmic}[1]
\Procedure{Offline-Optimisation}{}
\State $B = \mathcal{S} \times \mathcal{A}$  
\State $\theta_{\text{adv}} \gets \argmin_{\theta_{\text{adv}}} \text{MAE}\left(\pi_{\text{adv}}(B) - \hat{P}(B)\right)$  
\For {$i=1,2,\dots,I$} \Comment{Independent iterations}
\State Random initial state $s \sim P_0$.
\For { $t = 0,1,\dots,T-1$ } \Comment{Simulate trajectory}
\If{ $s$ is terminal}
\State \textbf{break}
\EndIf
\State Transition $(s,a \sim \pi(a \vert s),r(s,a),s' \sim \pi_{\text{adv}}(s' \vert s,a))$.
\State Policy gradient $\nabla_t \gets  \nabla_{\theta} \log\left(\pi_{\theta}(a\vert s)\right)$.
\State Adversary gradient $\nabla_t^{\text{adv}} \gets \nabla_{\theta_{\text{adv}}}\log\left(\pi_{\text{adv}}\right)$.
\State $s \gets s'$.
\EndFor
\State $T_{\text{stop}} \gets t$.
\For {$t = T_{\text{stop}}-1,T_{\text{stop}}-2,\dots,0$ }
\LineComment{Policy update}
\State $\mathbf{V}_{t} \gets V_t - \lambda C_t$.
\State $\theta \gets \theta + \eta_{1}(k) * \mathbf{V}_t \nabla_t$ 
\State $\lambda \gets \lambda + \eta_{2}(k) * (C - d)$ 
\LineComment{Adversary update}
\State $\nabla P \gets $\texttt{compute-nominal-deviation-grad}() 
\State $\theta_{\text{adv}} \gets \theta_{\text{adv}} - \eta_{1}(k)*\left(\mathbf{V}_{\text{next}} \nabla_t^{\text{adv}} + \lambda_{\text{adv}}\nabla P\right)$  
\State $\Delta P \gets ||\pi_{\text{adv}}(s_t,a_t) - \hat{P}_{s_t,a_t} ||_1$.
\State $\lambda_{\text{adv}} \gets \lambda_{\text{adv}} + \eta_{2}(k) * \left(\Delta P  - \alpha(s_t,a_t) \right)$ 
\EndFor
\EndFor
\State \textbf{return} $\pi_{\theta}$
\EndProcedure
\end{algorithmic}
\end{minipage}
\end{algorithm}

\begin{algorithm}
\caption{Computing gradient for deviation from nominal.} \label{alg: deviation-from-nominal}
\begin{minipage}[c]{0.95 \textwidth}
\small
\begin{algorithmic}[1]
\Procedure{Compute-nominal-deviation-grad}{ \newline parameters $\theta_{\text{adv}}$}
\LineComment{Set batch with small number of samples ($N_{\text{samp}}$)}
\State $B \gets \texttt{random}(\mathcal{S}\times\mathcal{A},N_{\text{samp}})$ 
\For {$i \in 1,\dots,B$}
\State $\alpha[i] \gets \alpha(s[i],a[i]) $.
\State $\text{nom}[i] \gets \hat{P}(s[i],a[i]) $
\State $y[i] \gets \pi_{\text{adv}}(: \vert s[i],a[i])$.
\State $\text{dev}[i] = \max\left(0,||y[i]  - \text{nom}[i] ||_1 - \alpha[i] \right)$
\EndFor
\State \textbf{return} $\nabla_{\theta_{\text{adv}}} \texttt{mean}(\text{dev})$
\EndProcedure
\end{algorithmic}
\end{minipage}
\end{algorithm}

\subsection{Uncertainty Set and Uncertainty Budget}
Uncertainty sets can be constructed in many ways. Methods of choice include sets based on Hoeffding's inequality and Bayesian methods \cite{Russel2019}. The experiments select Hoeffding L1 uncertainty sets which have state-action dependent transitions according to 
\begin{equation}
\mathcal{P}_{s,a} = \{ P :  ||P - \hat{P}_{s,a}|| < \alpha(s,a)  \} \,,
\end{equation}
where $\hat{P}$ is the nominal model, and $\alpha(s,a)$ is the uncertainty budget for state-action pair $(s,a)$. The uncertainty budget is set 
according to $\alpha(s,a) = \sqrt{\frac{2}{n(s,a)} \ln\left(\frac{2^S S A}{\delta})\right)}$, where $1 - \delta$ is the confidence and $n(s,a)$ 
 is the number of visitations of $(s,a)$. If $P^*$ is the true transition dynamics model, then the Hoeffding set ensures that $P^*_{s,a} \in 
\mathcal{P}_{s,a}$ with probability at least  $1 - \frac{\delta}{SA}$ and by union bound, that $P^* \in \mathcal{P}$ with probability at least $1 - \delta$ \cite{Russel2019}. 

\section{Lagrangian Policy Gradient Theorems}
\label{sec: theorems}
To prove that the desired objectives are indeed being optimised by Adversarial RCPG, two Lagrangian policy gradient theorems are derived. The first theorem shows that the Lagrangian of the policy can indeed be maximised using a simple policy gradient. The second theorem shows that the chosen adversary indeed follows the gradient steps to \textit{minimise} the Lagrangian of the policy. 
\subsection{Deriving the Robust Lagrangian Policy Gradient}
\label{sec: Robust Lagrangian Policy Gradient}
The key realisation for optimising the Lagrangian is that in CMDPs, both the value and the constraint-cost are expected cumulative quantities with the same discounting factor. This allows reusing existing results by reformulating the Lagrangian in terms of rewards and constraint-costs.
\begin{theorem}
\label{th: policy grad}
\textbf{Lagrangian policy gradient theorem.} 
Let $\pi: \mathcal{S} \to \Delta^A$ be a stochastic policy, let $P$ be the transition dynamics, let $s_0$ be the starting state, and for any state-action pair $(s,a) \in \mathcal{S} \times \mathcal{A}$ define $\mathbf{Q}_{\pi}(s,a) = Q_{\pi}(s,a) - \lambda C_{\pi}(s,a)$ and  $\mathbf{V}_{\pi}(s) = \mathbb{E}_{a \sim \pi(\cdot \vert s)}\left[ \mathbf{Q}_{\pi}(s,a) \right]$. Then it follows that 
\begin{equation}
\nabla_{\theta} \mathbf{V}_{\pi}(s_0) \propto \mathbb{E}_{\pi, P}\left[ \mathbf{Q}_{\pi}(s_t,a_t)  \nabla_{\theta}  \log\left(\pi(a_t \vert s_t)\right)  \right] \,. \end{equation}
\end{theorem}
\textbf{Proof:} The proof (see Appendix~A) reformulates the CMDP as a Lagrangian MDP \cite{Taleghan2018} and then makes analogous steps to the proof of the policy gradient theorem \cite{Sutton2017}.

As a consequence of this theorem,  updating with steps according to $\mathbf{Q}_{\pi}(s_t,a_t)  \nabla_{\theta}  \log\left(\pi(a_t \vert s_t)\right)$ will follow the gradient of the Lagrangian $L = V_{\pi}(s_0) - \lambda (C_{\pi}(s_0) - d)$ since the term $\lambda d$ is a constant. Since $P$ is arbitrarily chosen, this also holds for $P^+$ as defined in Eq.~\ref{eq: robustlagrangian}. 

\subsection{Deriving the Lagrangian Adversarial Policy Gradient}
\begin{theorem}
\label{th: adversarial grad}
\textbf{Lagrangian adversarial policy gradient theorem.} 
Let $\pi_{\text{adv}}: \mathcal{S}\times \mathcal{A} \to \Delta^S$ be the adversary replacing the transition dynamics of the CMDP, let $s_0$ be the starting state, let $T$ be the horizon of the decision process, and for any state-action pair $(s,a) \in \mathcal{S} \times \mathcal{A}$ define $\mathbf{Q}_{\pi}(s,a) = Q_{\pi}(s,a) - \lambda C_{\pi}(s,a)$ and $\mathbf{V}_{\pi}(s) = \mathbb{E}_{a \sim \pi(\cdot \vert s)}\left[ \mathbf{Q}_{\pi}(s,a) \right]$. Then it follows that
\begin{equation}
\nabla_{\theta_{\text{adv}}} \mathbf{V}_{\pi}(s_0) =\sum_{k=0}^{T-1}  \mathbb{E}\left[ \mathbf{V}_{\pi}(s_{k+1})  \nabla_{\theta_{\text{adv}}} \log\left( \pi_{\text{adv}}(s_{k+1} \vert s_{k},a_{k}) \right)  \right] \,.
\end{equation}
\end{theorem}
\textbf{Proof:} The proof uses a formalism similar to the previous proof but this time expands the gradient with respect to $\theta_{\text{adv}}$. The full proof is given in Appendix~B.

This theorem implies that applying consecutive updates of $  \mathbf{V}_{\pi}(s_{t+1})  \nabla_{\theta_{\text{adv}}} \log\left(\pi_{\text{adv}}(s_{t+1} \vert s_{t},a_t)\right) $ for $t=0,\dots, T-1$ will move $ \pi_{\text{adv}}$ along the gradient of the objective. 

\begin{table*}
\centering
\caption{\small{Comparison of algorithms on all tests. To provide a single statistic, the penalised return \cite{Mankowitz2020a} is defined as $R_{\text{pen}} = V(s_0) - \bar{\lambda} \max\left(0,\left( C(s_0) - d \right)\right)$. The evaluation weight is set as $\bar{\lambda}=500$, which is equal to the maximal Lagrangian multiplier during the constrained optimisation. Bold highlights the top two scores while underline indicates the highest score.}}\label{tab: Rpenalised}
\resizebox{16cm}{!}{
\begin{tabular}{l p{2.2cm} p{2.2cm} p{2.2cm} p{2.2cm} p{2.2cm} p{2.2cm}}
\toprule
& Adversarial RCPG & RCPG \newline (Robust Lagrangian) & RCPG \newline (Robust value) & RCPG \newline  (Robust constraint) & CPG & PG \\
\midrule
Inventory Management & $\mathbf{-3058.6 \pm 1341.6}$ & $\underline{\mathbf{31.8 \pm 26.3}}$ & $ -3843.1 \pm 1567.9$ & $ -7812.0 \pm 2148.4$ & $ -3977.0 \pm 1243.6$ & $ -28092.3 \pm 913.1$ \\
Safe Navigation 1A &   $\mathbf{\underline{-76.7 \pm 20.2}}$ & $ -190.5 \pm 9.2$&   $ -4825.8 \pm 4537.9$ & $ -200.0 \pm 0.0$&  $\mathbf{-133.6 \pm 20.2}$&  $ -9443.7 \pm 6307.1$ \\
Safe Navigation 1B &  $\mathbf{\underline{-71.9 \pm 18.9}}$  & $ -273.9 \pm 81.7$  & $ -4751.3 \pm 3965.5$ & $-735.3 \pm 312.2$ & $\mathbf{-123.5 \pm 18.7}$ & $-8686.0 \pm 5718.3$ \\
Safe Navigation 2A & $ \mathbf{-48.1 \pm 9.7}$&  $ -316.8 \pm 282.0$  & $ -275.7 \pm 224.0$ & $\mathbf{\underline{-30.6 \pm 8.1}}$  & $ -259.1 \pm 218.3$  & $ -512.4 \pm 299.0$\\
Safe Navigation 2B  & $\mathbf{-1437.2 \pm 107.4}$ & $ -1451.0 \pm 221.9$& $ -1825.0 \pm 391.8$& $\mathbf{\underline{-1259.3 \pm 101.2}}$& $ -1681.6 \pm 421.2$ & $ -2395.5 \pm 546.5$\\
\bottomrule
\end{tabular}
}
\end{table*}

\section{Results}
\label{sec: experiments}
Having introduced Adversarial RCPG and RCPG, the experimental validation below compares their performance on the cumulative reward and constraint-cost in perturbed environments. The experiments are set up in three consecutive phases. In the \textbf{model estimation phase}, a random uniform policy is run on a dynamics model $P_{\text{data}}$, which represents a centroid of the test dynamics models. The result of phase 1 is a nominal model $\hat{P}$ and (if applicable) the Hoeffding L1 uncertainty set $\mathcal{P}$. In the \textbf{policy training phase}, policies are trained across 5,000 episodes based on either $P_{\text{data}}$ (non-robust algorithms) or $\mathcal{P}$. In the  \textbf{policy test phase}, the trained policy is tested by taking greedy actions on a set of test dynamics that are perturbations of $P_{\text{data}}$. To evaluate the training and test performance, the value and constraint-cost are evaluated without discounts, and the resulting budget $d$ is corrected accordingly by a factor $T/(\sum_{i=0}^{T-1} \gamma^i)$, where $T$ is the maximal episode length.

The algorithms evaluated are the following: 1) the \textbf{Adversarial RCPG} algorithm implementing the Lagrangian adversary as described in Algorithm~\ref{alg: Adversarial RCPG} and supported by Theorem~\ref{th: policy grad}--\ref{th: adversarial grad}; 2) \textbf{RCPG (Robust Lagrangian)}, the variant of RCPG formulated in Section~\ref{sec: Lagrangian RCPG} to formulate the worst-case dynamics as the model that minimises the Lagrangian, as supported by Theorem~\ref{th: policy grad}, which can be seen as an ablation that removes the adversary but keeps the Lagrangian objective; 3) \textbf{RCPG (Robust value)} \cite{Russel2021}, which formulates robustness in terms of the dynamics with worst-case value; 4) \textbf{RCPG (Robust constraint)}, which formulates robustness in terms of the dynamics with worst-case constraint-cost \cite{Russel2021,Bossens2022}; 5) \textbf{CPG}, which uses the nominal transition dynamics instead of the worst-case transition dynamics, as an ablation without robustness; and 6) \textbf{PG}, a further ablation condition with no constraints, which corresponds to REINFORCE \cite{Sutton2017}.

To demonstrate a range of applications, experiments include an inventory management domain and two safe navigation tasks, each with a variety of test cases. As a quick overview of the test results, Tab.~\ref{tab: Rpenalised} shows that the Adversarial RCPG is always among the top two performing algorithms on the penalised return, a performance metric for CMDPs. The reader may also refer to Appendix C, D, and E of the supplementary information for additional details and figures of the experiments. The source code used for the experiments is available at \url{https://github.com/bossdm/RCMDP}.

\subsection{Inventory Management}
\begin{figure}[htbp!]
\centering
\includegraphics[width=0.40\textwidth]{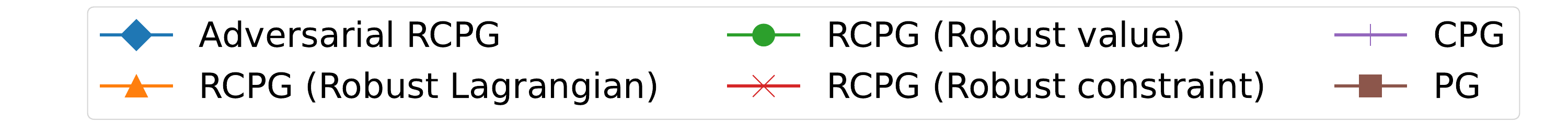} \\
\subfloat[Test value]{
\includegraphics[width=0.22\textwidth]{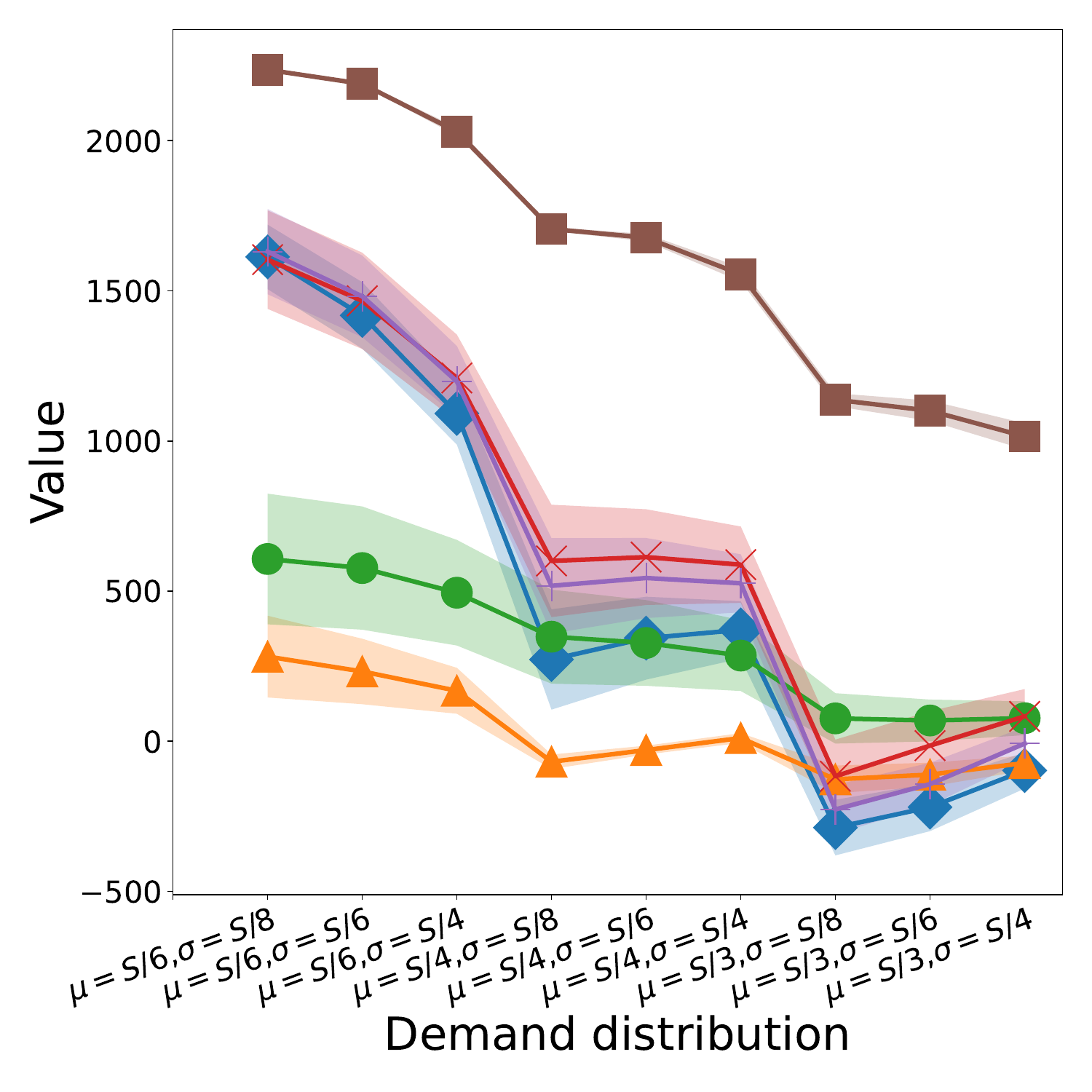}
}
\subfloat[Test overshoot]{\includegraphics[width=0.22\textwidth]{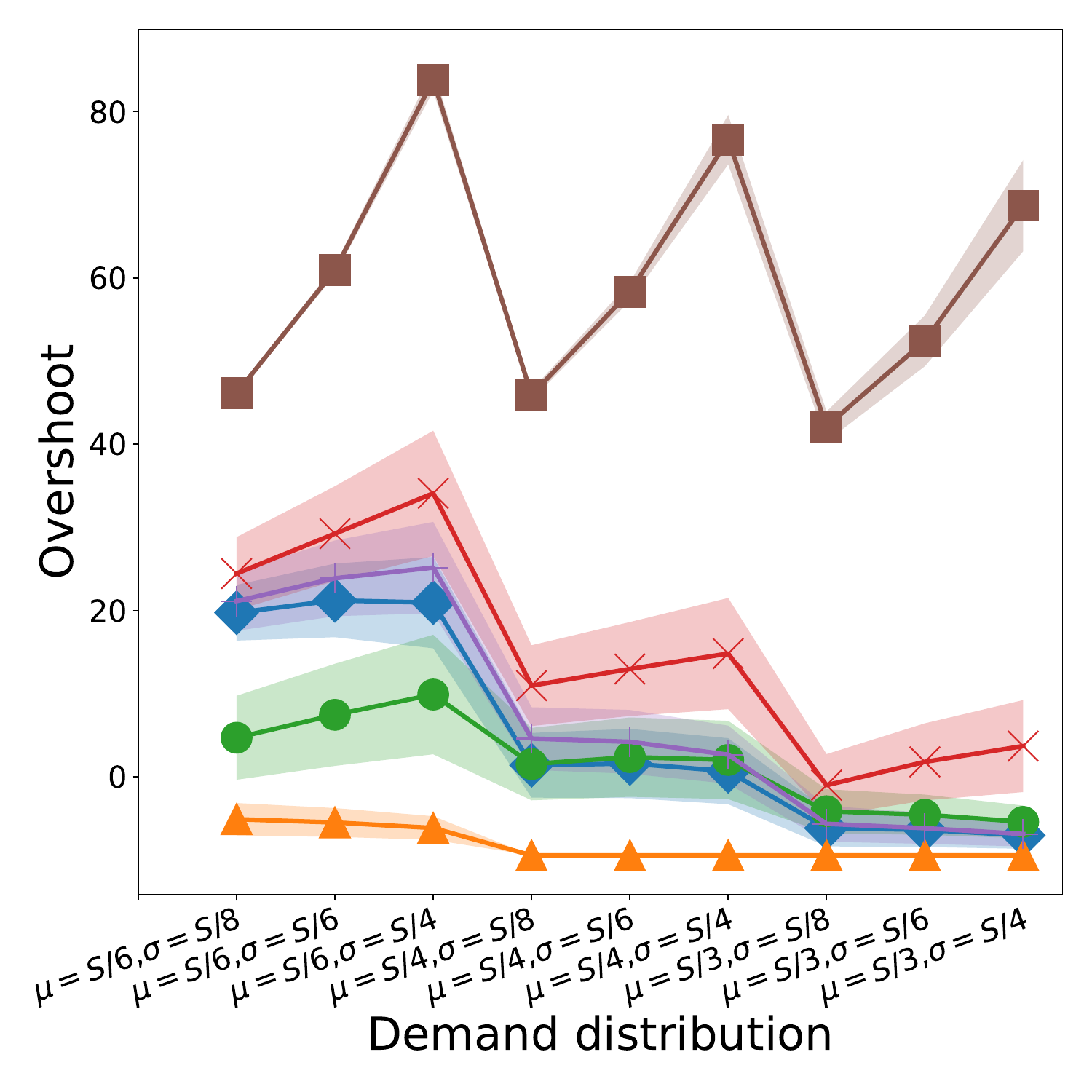}}
\caption{Test performance metrics of the algorithms on the test set of perturbed transition dynamics in Inventory Management. For each of 20 training runs, each parameter setting is run 50 times and the plot displays the mean and standard error over runs. The parameter manipulated is the mean, $\mu$, and standard deviation, $\sigma$, of the demand distribution.} \label{fig: test-IM}
\end{figure}

The first domain is the inventory management problem \cite{Puterman2005}, which has been the test bed of the RCPG algorithm 
\cite{Russel2021}. The task of the agent is to purchase items to make optimal profits selling the items, 
balancing supply with demand in the process. The state is the current inventory while the action is the purchased number of items from the supplier. States are integers in $\{0,1,\dots,S-1\}$, where $S$ is the number of states. Initially the inventory is empty, corresponding to initial state $s_0 = 0$, and a full inventory contains $S-1$ items. The constraint is that the number of purchased items, $a \in \mathcal{A}$, should not exceed the purchasing limit. Oscillating behaviours shown by RCPG algorithms (see Appendix~E) are attributed to large abrupt changes in the estimated worst-case distribution. The policy test 
phase consists of 9 different parameters $\mu$ and $\sigma$ for the demand distribution, resulting in changed transition dynamics. The penalised return scores in Tab.~\ref{tab: Rpenalised} demonstrates that RCPG (Robust Lagrangian) has the highest penalised return. This indicates that robustifying the Lagrangian is beneficial but also that the abrupt changes in the estimated worst-case distribution do not appear to hamper RCPG's test performance;  because any state is reachable from any other state and the demand is \textit{iid}, the algorithm is less sensitive to excessive sampling of a single state and shifting transition dynamics. Fig.~\ref{fig: test-IM} demonstrates the value, which is proportional to the profit, and the overshoot across the different demand distributions.

\subsection{Safe Navigation}
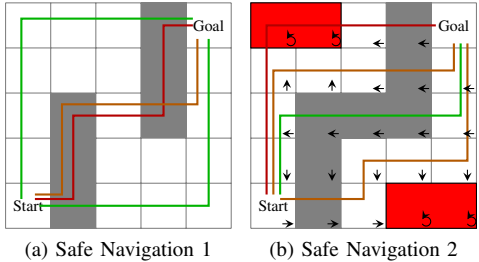
\begin{figure}[htbp!]
\centering
\subfloat[Safe Navigation 1]{
\begin{tikzpicture}[scale=0.60,every node/.style={scale=0.60}]
\draw[step=1.0cm,color=gray] (0,0) grid (5,5);
\path [fill=gray] (1,0) rectangle (2,3);
\path [fill=gray] (3,2) rectangle (4,5);
\node at (0.5,0.5) {Start};
\node at (4.5,4.5) {Goal};
\draw[thick,black!30!red] (0.65,0.65) -- (1.5,0.65) -- (1.5,2.5) -- (1.5,2.5) -- (3.5,2.5) -- (3.5,2.5) -- (3.5,4.5) -- (4.15,4.5);
\draw[thick,black!30!orange] (0.65,0.75) -- (1.25,0.75) -- (1.25,2.75) --  (4.25,2.75) -- (4.25,4.2);
\draw[thick,black!30!green] (0.75,0.5) -- (4.5,0.5) -- (4.5,4.20);
\draw[thick,black!30!green] (0.35,0.65) -- (0.35,4.65) -- (4.15,4.65);
\end{tikzpicture}
}
\subfloat[Safe Navigation 2]{
\begin{tikzpicture}[scale=0.60,every node/.style={scale=0.60}]
\draw[step=1.0cm,color=gray] (0,0) grid (5,5);
\path [fill=gray] (1,0) rectangle (2,3);
\path [fill=gray] (2,2) rectangle (3,3);
\path [fill=gray] (3,2) rectangle (4,5);
\draw [fill=red] (0,4) rectangle (2,5);
\draw [fill=red] (3,0) rectangle (5,1);
\node at (0.5,0.5) {Start};
\node at (4.5,4.5) {Goal};
\draw[thick,black!30!red] (0.35,0.75) -- (0.35,4.5) -- (4.10,4.5);
\draw[thick,black!30!orange] (0.5,0.75) -- (0.5,3.5) -- (4.5,3.5) -- (4.5,4.1);
\draw[thick,black!30!orange] (0.65,0.65) -- (2.5,0.65) -- (2.5,1.5) -- (4.8,1.5) -- (4.8,4.1);
\draw[thick,black!30!green] (0.65,0.75) -- (0.65,2.5) -- (4.65,2.5) -- (4.65,4.1);
% y== 0 (towards red)
\draw[->,-stealth] (0.70,0.1) -- (0.95,0.1); 
\draw[->,-stealth] (1.70,0.1) -- (1.95,0.1); 
\draw[->,-stealth] (2.70,0.1) -- (2.95,0.1); 
\draw[->,-stealth] (3.80,0.1) arc(-150:150:0.10); 
\draw[->,-stealth] (4.80,0.1) arc(-150:150:0.10); 
% y==1 (down)
\draw[->,-stealth] (0.80,1.3) -- (0.8,1.05); 
\draw[->,-stealth] (1.80,1.3) -- (1.8,1.05); 
\draw[->,-stealth] (2.80,1.3) -- (2.8,1.05); 
\draw[->,-stealth] (3.80,1.3) -- (3.8,1.05); 
\draw[->,-stealth] (4.80,1.3) -- (4.8,1.05); 
% y==2 (back)
\draw[->,-stealth] (0.95,2.1) -- (0.70,2.1); 
\draw[->,-stealth] (1.95,2.1) -- (1.70,2.1); 
\draw[->,-stealth] (2.95,2.1) -- (2.70,2.1); 
\draw[->,-stealth] (3.95,2.1) -- (3.70,2.1); 
\draw[->,-stealth] (4.95,2.1) -- (4.70,2.1); 
% y==3 (up and back)
\draw[->,-stealth] (0.8,3.05) -- (0.80,3.3); 
\draw[->,-stealth] (1.8,3.05) -- (1.80,3.3); 
\draw[->,-stealth] (2.95,3.1) -- (2.70,3.1); 
\draw[->,-stealth] (3.95,3.1) -- (3.70,3.1); 
\draw[->,-stealth] (4.95,3.1) -- (4.70,3.1); 
% y==4 (towards red)
\draw[->,-stealth] (0.80,4.1) arc(-150:150:0.10); 
\draw[->,-stealth] (1.80,4.1) arc(-150:150:0.10); 
\draw[->,-stealth] (2.95,4.1) -- (2.70,4.1); 
\draw[->,-stealth] (3.95,4.1) -- (3.70,4.1); 

\end{tikzpicture}
}
\caption{Illustration of the safe navigation tasks. In Safe Navigation 1, the constraint is to hit no more than 4 grey cells on average. In Safe Navigation 2, the constraint is to avoid the red cells and only a limited number of grey cells. Unconstrained solutions, constrained, and robust-constrained trajectories are demonstrated in red, orange, and green, respectively. The arrows represent the worst-case transitions for test Safe Navigation 2B.}\label{fig: maze}
\end{figure}

The second domain and third domain are safe navigation tasks in a 5-by-5 square grid world, formulated specifically to highlight the advantages of 
agents that satisfy constraints robustly (see Fig.~\ref{fig: maze}). 

\begin{figure}[htbp!]
\centering
\includegraphics[width=0.40\textwidth]{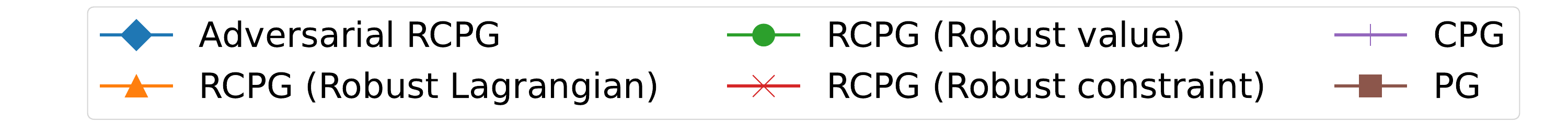} \\
\subfloat[Test A value]{
\includegraphics[width=0.22\textwidth]{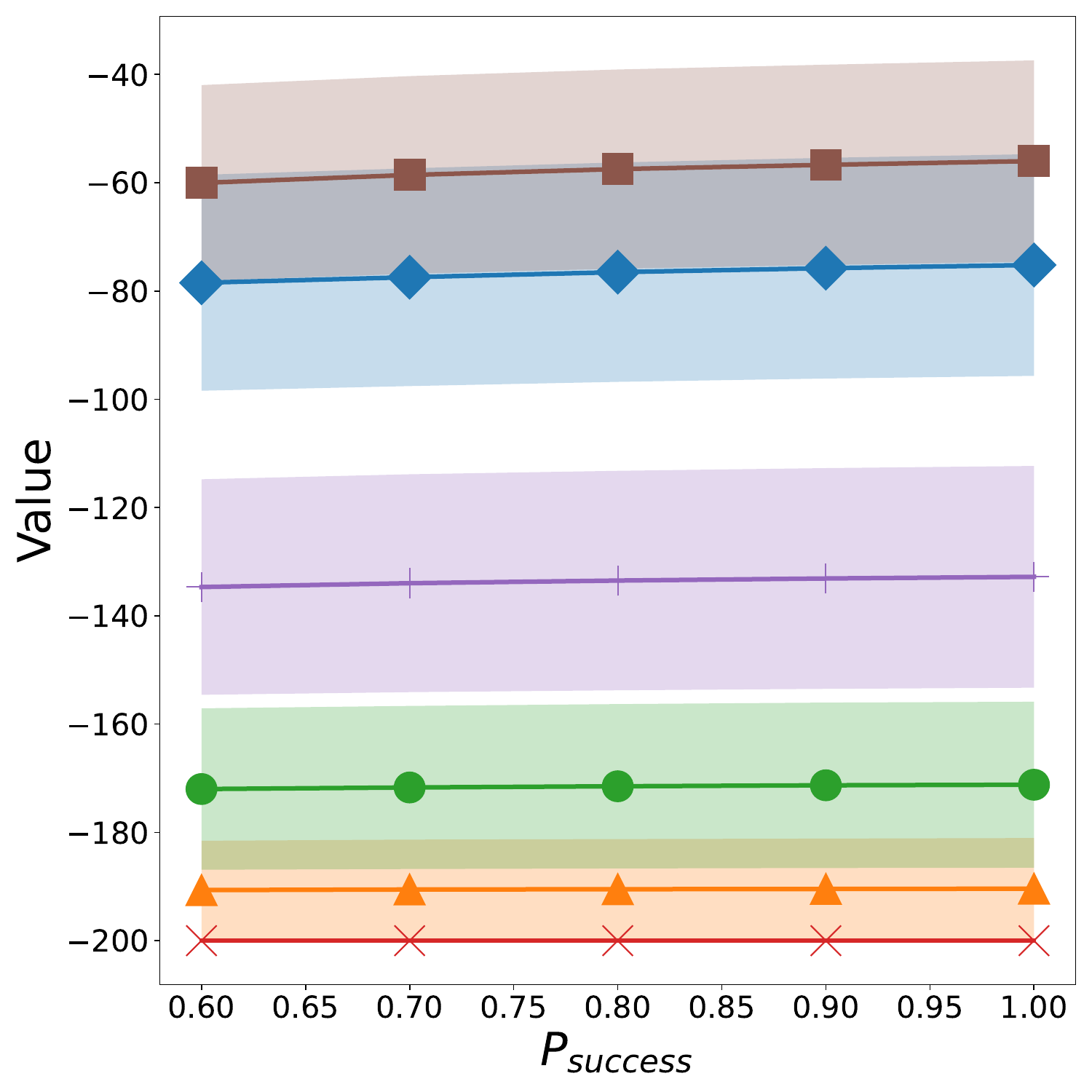}
}
\subfloat[Test A overshoot]{
\includegraphics[width=0.22\textwidth]{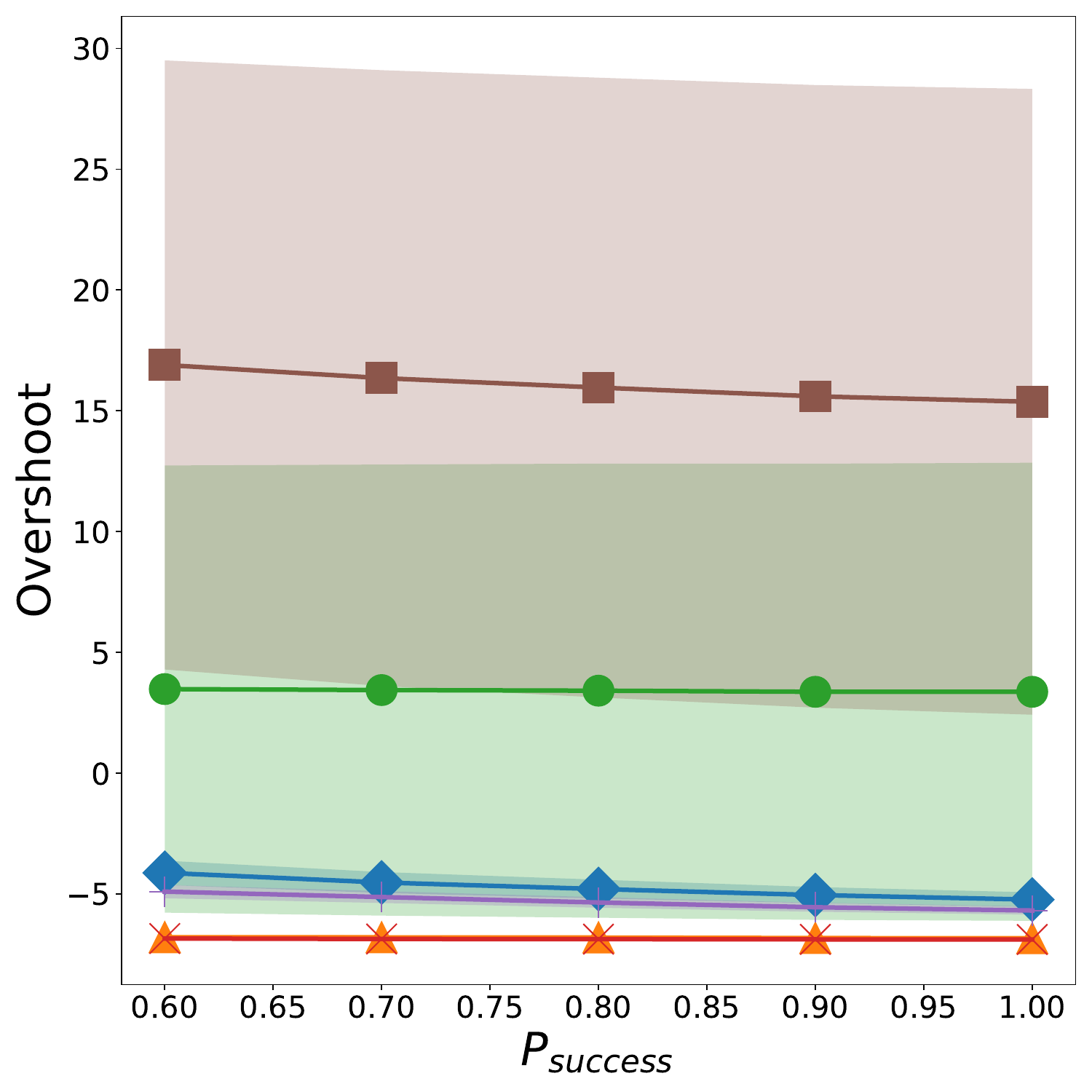}
} \\
\subfloat[Test B value]{
\includegraphics[width=0.22\textwidth]{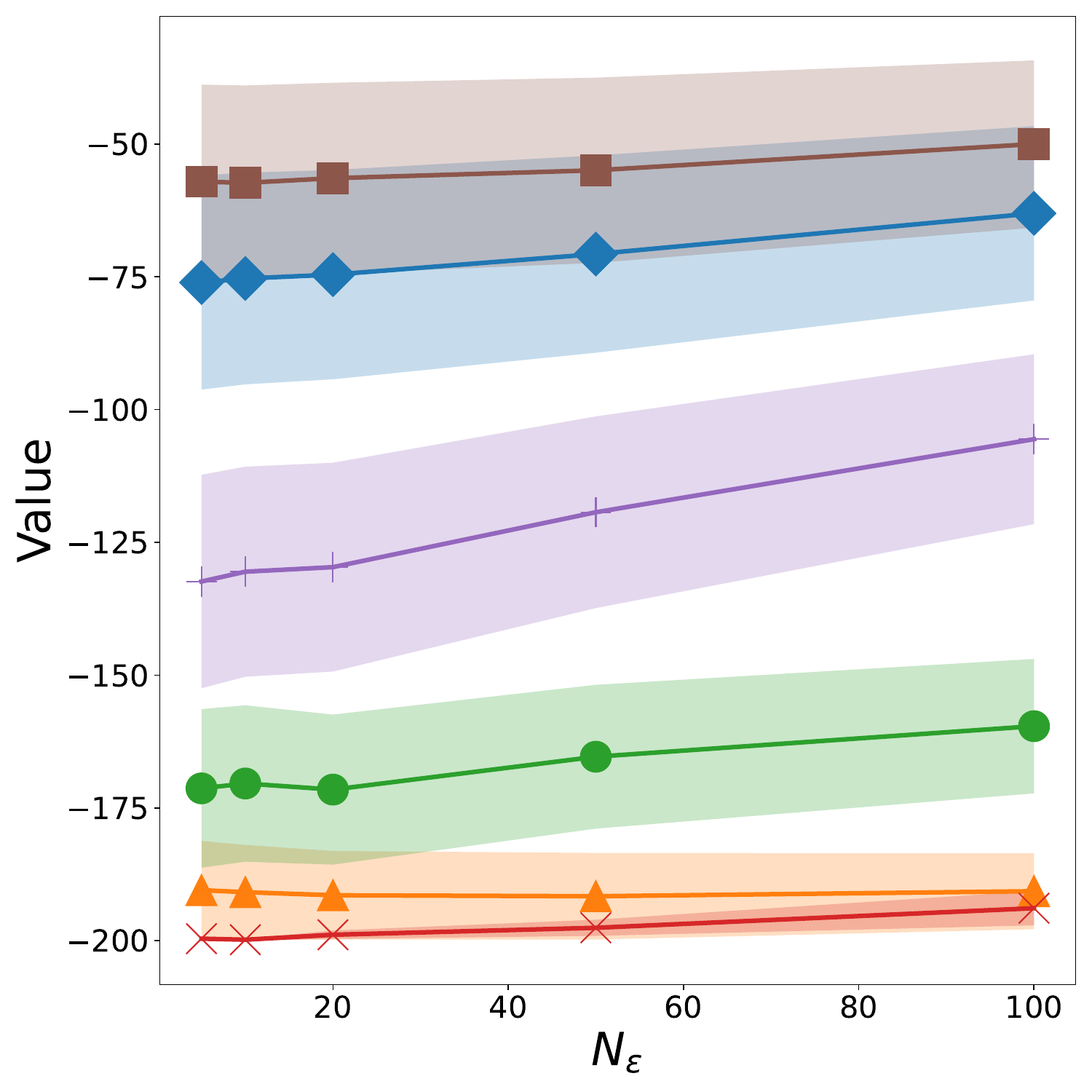}
}
\subfloat[Test B overshoot]{
\includegraphics[width=0.22\textwidth]{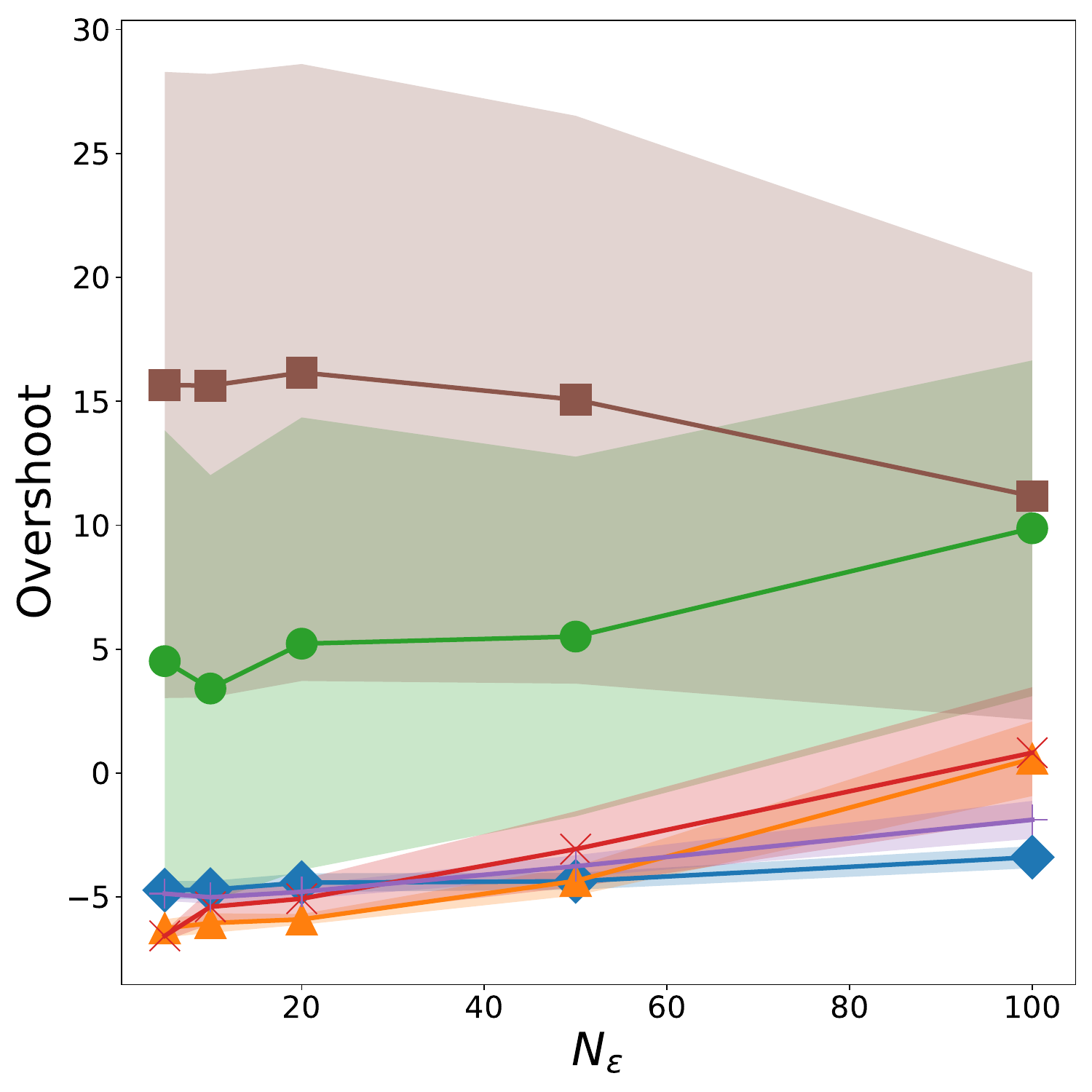}
}
\caption{Test performance metrics of the algorithms on the test set of perturbed transition dynamics in Safe Navigation 1. For each of 20 training runs, each parameter setting is run 50 times and the plot displays the mean and standard error over runs. \textbf{Test A:} The parameter manipulated is the move probability of the actions. \textbf{Test B:} The parameter manipulated is the number of perturbations,   i.e. randomly selected state-action pairs that are perturbed with a random offset in $\mathcal{N}(s)$.} \label{fig: SafeNavigation1 tests}
\end{figure}

\paragraph{Safe Navigation 1}
In Safe Navigation 1 (see Fig.~\ref{fig: maze}a), the grid contains 6 grey cells that incur constraint-cost of $1.0$ and agents should satisfy a budget of $d = 3.0$. Agents stay in the grid world for $T=200$ time steps if the goal is not found. Test 1A manipulates $P_{\text{success}}$, the probability with which the agent successfully moves to the intended location. Test 1B fixes $P_{\text{success}}=0.80$ while manipulating $N_{\epsilon}$, the number of state-action pairs perturbed by setting $s' = s+\epsilon(s,a)$ upon successful action. As shown in  Tab.~\ref{tab: Rpenalised}, Adversarial RCPG outperforms all other algorithms in both tests of Safe Navigation 1. In line with the hypothesis that Adversarial RCPG provides incremental learning, the training value and constraint-overshoot develop much more smoothly in Adversarial RCPG when compared to RCPG variants, which display oscillating and high-variance scores on these metrics (see Appendix~E). In the test, the RCPG algorithms do not find paths to the goal location although the RCPG (Robust constraint) and RCPG (Robust Lagrangian) satisfy the constraint. As shown in Fig.~\ref{fig: SafeNavigation1 tests}, Adversarial RCPG combines a high value comparable to PG with a negative overshoot that is not affected even by severe perturbations. Safe Navigation 1 presents a particular challenge for RCPG; this is attributed to learning paths that satisfy the constraint (by avoiding grey cells) but that do not come closer to the goal.

\begin{figure}[htbp!]
\centering
\includegraphics[width=0.44\textwidth]{figures/legend.pdf} \\
\subfloat[Test A value]{
\includegraphics[width=0.22\textwidth]{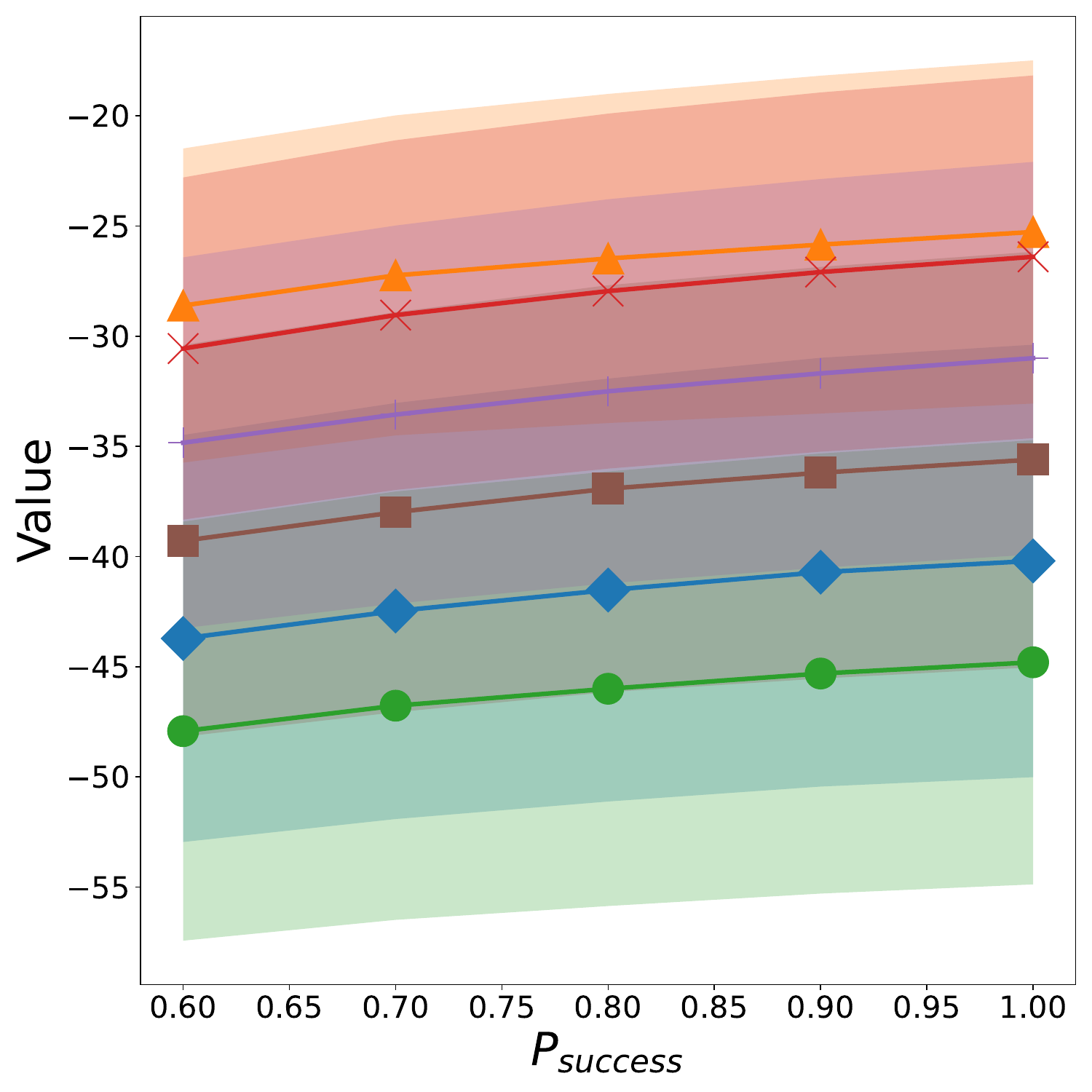}
}
\subfloat[Test A overshoot]{
\includegraphics[width=0.22\textwidth]{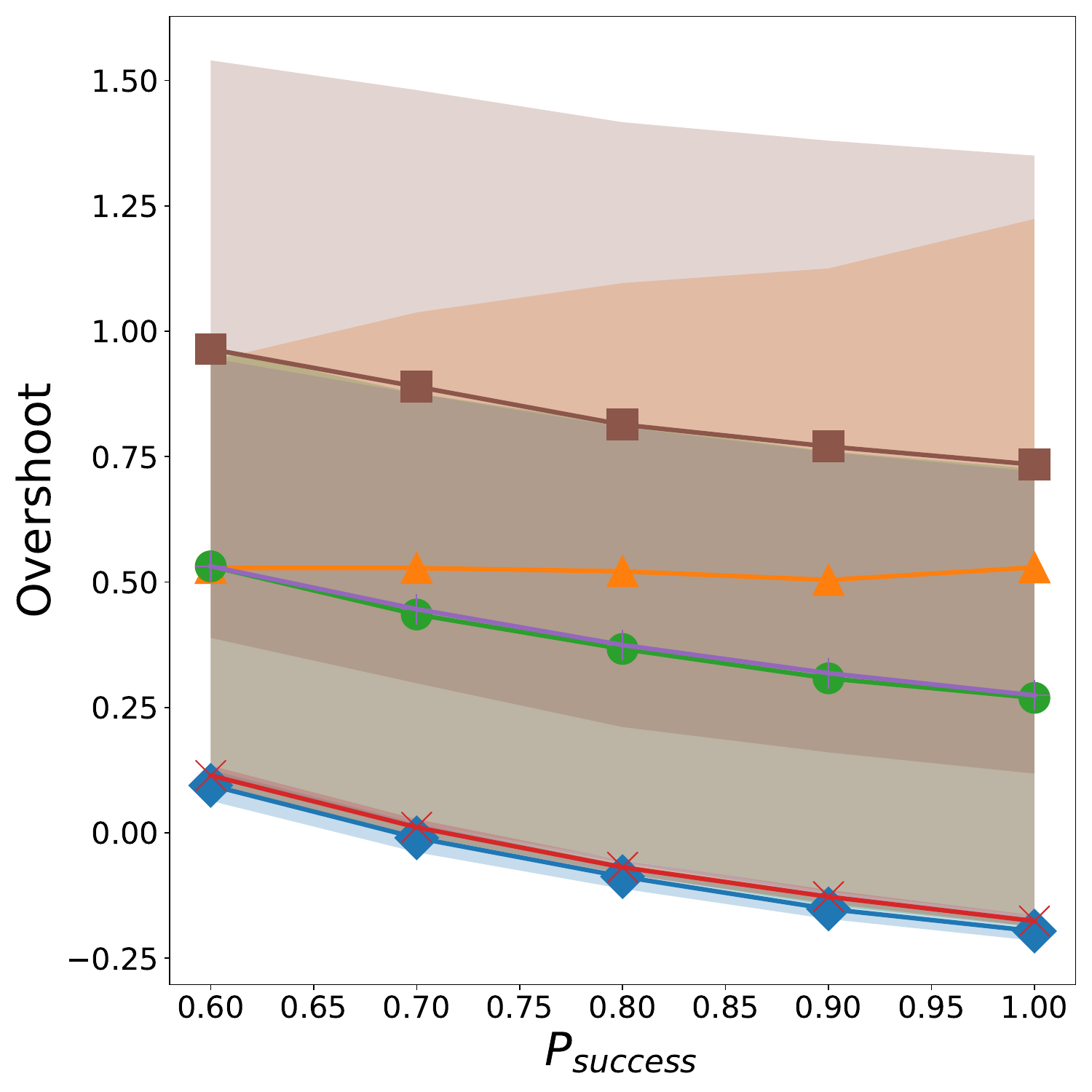}
}\\
\subfloat[Test B value]{
\includegraphics[width=0.22\textwidth]{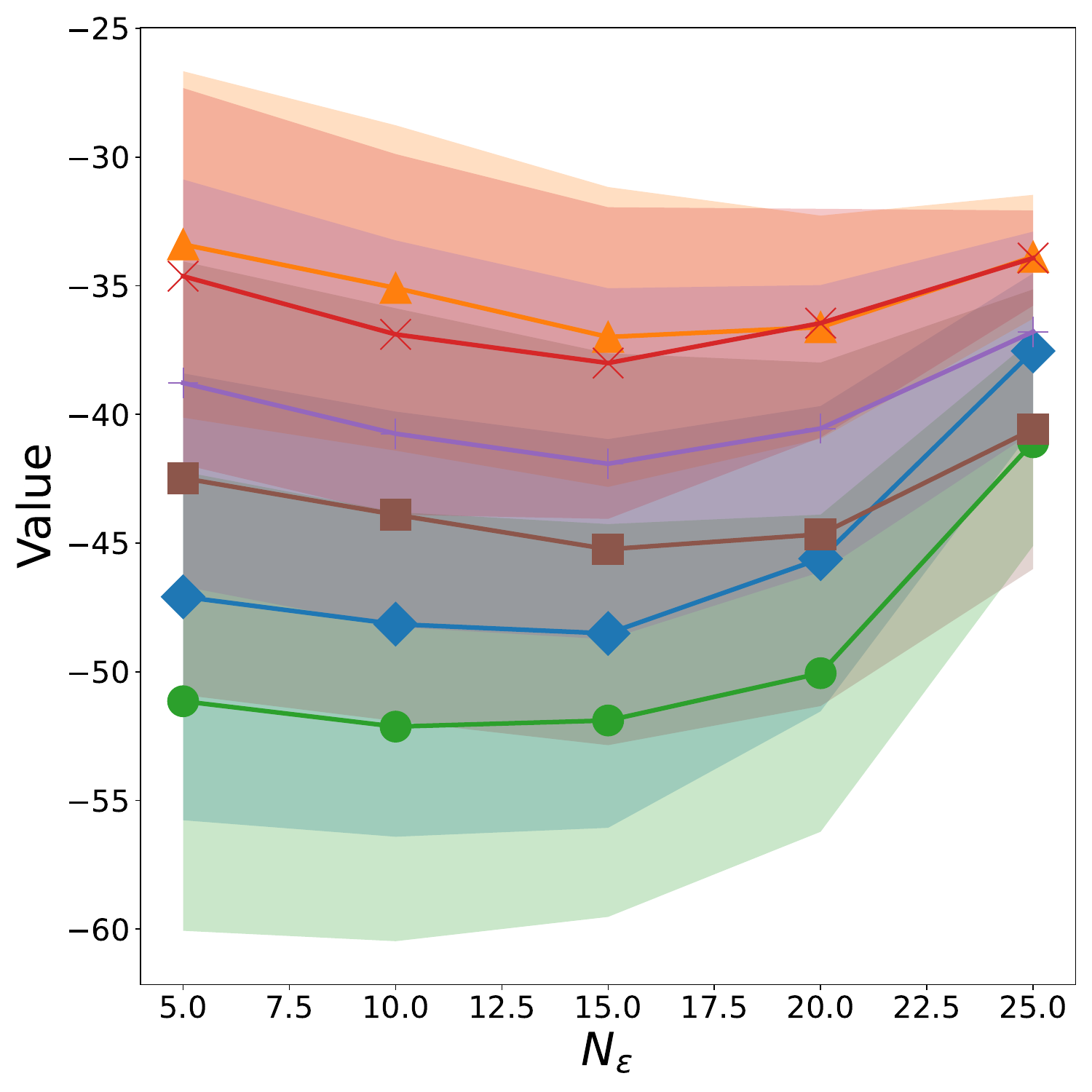}
}
\subfloat[Test B overshoot]{\includegraphics[width=0.22\textwidth]{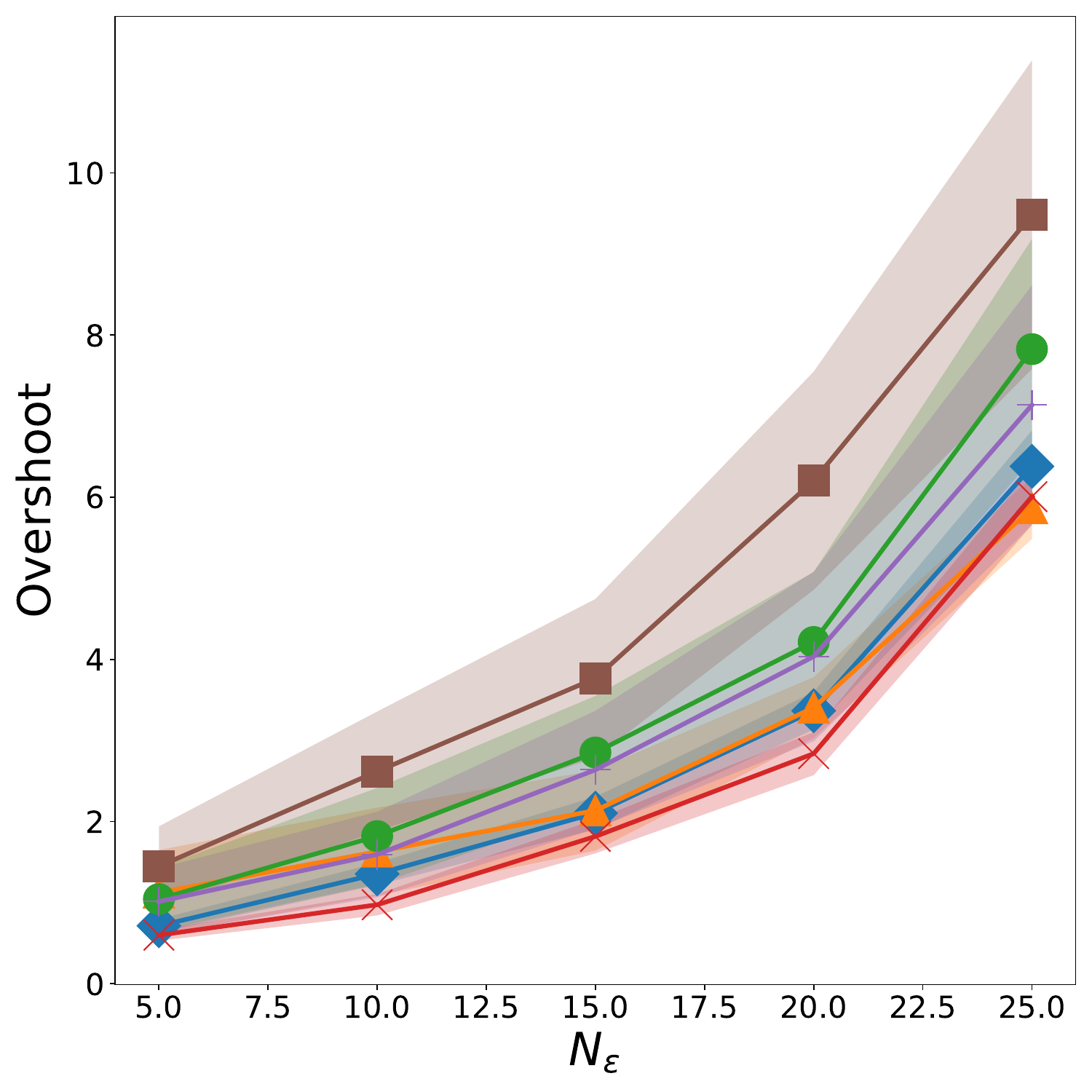}
}
\caption{Test performance metrics of the algorithms on the test set of perturbed transition dynamics in Safe Navigation 2. For each of 20 training runs, each parameter setting is run 50 times and the plot displays the mean and standard error over runs. \textbf{Test A:} The parameter manipulated is the move probability of the actions. \textbf{Test B:} The parameter manipulated is the number of perturbations, i.e. randomly selected states that are perturbed with a worst-case transition according to the arrows in Fig.~\ref{fig: maze}b.} \label{fig: SafeNavigation2 tests}
\end{figure}

\paragraph{Safe Navigation 2}
In Safe Navigation 2 (see Fig.~\ref{fig: maze}b), the grid contains 7 grey cells that incur constraint-cost of $0.1$, 4 red cells that incur a 
cost of $1.0$, and agents should satisfy a budget of $d = 0.4$. Agents stay in the grid world for $T=100$ time steps if the goal is not found.  Test 2A manipulates $P_{\text{success}}$. Test 2B keeps $P_{\text{success}}=0.50$ and upon failure, the agent is moved according to worst-case transitions as shown in the arrows of Fig.~\ref{fig: maze}b. As shown in  Tab.~\ref{tab: Rpenalised}, RCPG (Robust constraint) is the top performer followed by Adversarial RCPG in both tests of Safe 
Navigation 2. The tighter uncertainty set and shorter episode leads to a similarly smooth training for all RCPG variants when compared to Adversarial RCPG (see Appendix~E) as it makes starting from the worst-case dynamics less challenging. Avoiding constraint-cost becomes comparably more challenging than eventually finding the goal; therefore it becomes more important to robustify the constraint-cost compared to the value (see highest rank of RCPG with Robust constraint). As shown in Fig.~\ref{fig: SafeNavigation2 tests}, Adversarial RCPG does not have the highest value but achieves a low overshoot comparable to RCPG (Robust constraint); RCPG (Robust Lagrangian) also performs comparably on the overshoot on test B.

\section{Conclusion and future work}
Providing robustness as well as constraints into policies is of critical importance for safe reinforcement learning. This paper proposes a robust Lagrangian objective and an adversarial gradient descent algorithm for a modified robust constrained policy gradient algorithm with stable and incremental learning properties. These modifications are empirically demonstrated to improve reward-based and constraint-based metrics on a wide range of test perturbations. Adversarial policies have been of interest in designing realistic attacks on reinforcement learning policies (e.g. \cite{Gleave2020,Mandlekar2017}); while these works do not consider uncertainty sets, an interesting avenue for future research is to extend Adversarial RCPG with uncertainty sets that satisfy realism constraints in addition to the current norm constraints.

\bibliographystyle{ieeetr}
\bibliography{library}

\clearpage
\onecolumn
{\huge{Supplementary Information}}
\section*{Appendix A: Proof of Robust Constrained Policy Gradient theorem}
\label{appendixA}
Using the notation $\mathbf{r}(s,a) = r(s,a) - \lambda c(s,a)$ to formulate the problem as an MDP, we have
\begin{align*}
\nabla_{\theta} \mathbf{V}_{\pi}(s) &=  \nabla_{\theta} \left( \sum_{a \in \mathcal{A}} \pi(a \vert s) \mathbf{Q}_{\pi}(s,a) \right) \quad \quad \text{(definition)} \\
						   &= \sum_{a \in \mathcal{A}} \mathbf{Q}_{\pi}(s,a) \nabla_{\theta} \pi(a \vert s)  + \pi(a \vert s) \nabla_{\theta}  \mathbf{Q}_{\pi}(s,a) \quad \quad \text{(product rule)} \\
						   						   &= \sum_{a \in \mathcal{A}} \mathbf{Q}_{\pi}(s,a) \nabla_{\theta} \pi(a \vert s)  + \pi(a \vert s)  \nabla_{\theta} \sum_{s',\mathbf{r}} \mathbb{P}(s', \mathbf{r} \vert a,s) \left( \mathbf{r}(s,a) + \mathbf{V}_{\pi}(s') \right) \quad \quad \text{(bootstrap from next Q)} \\
						   						   	   &= \sum_{a \in \mathcal{A}} \mathbf{Q}_{\pi}(s,a) \nabla_{\theta} \pi(a \vert s)  + \pi(a \vert s) \nabla_{\theta} \sum_{s'} P(s' \vert a,s) \nabla_{\theta} \mathbf{V}_{\pi}(s') \\ &  \quad \quad  \text{(noting that ($P(s' \vert a,s) = \sum_{r} \mathbb{P}(s', \mathbf{r} \vert a,s)$)} \\
						   						   	   &= \sum_{a \in \mathcal{A}} \mathbf{Q}_{\pi}(s,a)  \nabla_{\theta} \pi(a \vert s) + \pi(a \vert s) \sum_{s'} P(s' \vert s,a)  \\ & \quad \quad \left( \sum_{a' \in \mathcal{A}} \nabla_{\theta} \pi(a' \vert s') \mathbf{Q}_{\pi}(s',a') +   \pi(a' \vert s') \sum_{s''} P(s'' \vert s',a') \nabla_{\theta} \mathbf{V}_{\pi}(s'')\right)  \quad \quad \quad \text{(unpacking analogously)} \\
						   						   	   &= \sum_{s_{\text{next}} \in \mathcal{S}} \sum_{k=0}^{\infty} \mathbb{P}( s \to  s_{\text{next}}  \vert k, \pi)  \sum_{a} \mathbf{Q}_{\pi}(s_{\text{next}},a) \nabla_{\theta} \pi(a \vert s_{\text{next}})\,.  \quad \quad \quad \text{(repeated unpacking)} \\
\end{align*}

To demonstrate the objective is satisfied from $t=0$ to $t=\infty$, the proof continues from the initial state $s_0$. There it is useful to consider the average number of visitations of $s$ in an episode, $n(s) := \sum_{k=0}^{\infty} \mathbb{P}(s_0 \to s \vert k, \pi)$, and its relation to the on-policy distribution $\mu(s \vert \pi)$, the fraction of time spent in each state when taking actions from $\pi$:
\begin{align*}
\nabla_{\theta} \mathbf{V}_{\pi}(s_0) &= \sum_{s \in \mathcal{S}} n(s)  \sum_{a}  \mathbf{Q}_{\pi}(s,a) \nabla_{\theta} \pi(a \vert s) \\
									  &\propto \sum_{s \in \mathcal{S}} \mu(s \vert \pi)  \sum_{a} \mathbf{Q}_{\pi}(s,a) \nabla_{\theta} \pi(a \vert s)   \\
									  &= \mathbb{E}_{\pi, P}\left[ \sum_a \mathbf{Q}_{\pi}(s_t,a) \nabla_{\theta} \pi(a \vert s_t)   \right] \\
									  &= \mathbb{E}_{\pi, P}\left[ \sum_a \mathbf{Q}_{\pi}(s_t,a) \pi(a \vert s_t) \frac{\nabla_{\theta} \pi(a \vert s_t)}{\pi(a \vert s_t)}   \right]  \\
									   &= \mathbb{E}_{\pi, P}\left[\frac{\mathbf{Q}_{\pi}(s_t,a_t)  \nabla_{\theta} \pi(a_t \vert s_t)}{\pi(a_t \vert s_t)}   \right] \\
									  &= \mathbb{E}_{\pi, P}\left[\mathbf{Q}_{\pi}(s_t,a_t)  \nabla_{\theta}  \log\left(\pi(a_t \vert s_t)\right)  \right] \,.
\end{align*}
\qed
\newpage
\section*{Appendix B: Proof of Robust Constrained Adversarial Policy Gradient Theorem}
\label{appendixB}
First note that the gradient of $\mathbf{V}_{\pi}(s_t)$ of a state $s_t$ at time $t$ is given by
\begin{align*}
 \nabla_{\theta_{\text{adv}}} \mathbf{V}_{\pi}(s_t) &=  \nabla_{\theta_{\text{adv}}} \left( \sum_a \pi(a \vert s_t) \mathbf{Q}_{\pi}(s_t,a)  \right)   \quad \quad \text{(definition)} \\
 													&=   \sum_a \pi(a \vert s_t)  \nabla_{\theta_{\text{adv}}} \mathbf{Q}_{\pi}(s_t,a) \quad\quad  \text{($\pi$ independent of $\pi_{\text{adv}}$ )}\\
 													&= \sum_a \pi(a \vert s_t) \nabla_{\theta_{\text{adv}}}  \left( \mathbb{P}(s',\mathbf{r} \vert s_t, a) \left( \sum_{\mathbf{r},s'} \mathbf{r}(s_t,a) + \mathbf{V}_{\pi}(s') \right) \right)  \quad\quad \text{(expand the Q-value)} \\
 													&= \sum_a \pi(a \vert s_t)  \nabla_{\theta_{\text{adv}}}  \left( \sum_{s'} \mathbb{P}(s' \vert s_{t},a) \mathbf{V}_{\pi}(s') \right) \quad\quad   \text{(reward distribution independent of $\pi_{\text{adv}}$)} \\
 													&= \sum_a \pi(a \vert s_t)  \nabla_{\theta_{\text{adv}}} \left( \sum_{s'} \pi_{\text{adv}}(s' \vert s_{t},a) \mathbf{V}_{\pi}(s') \right) \quad\quad   \text{(use $\pi_{\text{adv}}$ to generate the next state)} \\
 													&= \sum_a \pi(a \vert s_t)  \left( \sum_{s'} \mathbf{V}_{\pi}(s') \nabla_{\theta_{\text{adv}}}\pi_{\text{adv}}(s' \vert s_{t},a) +  \pi_{\text{adv}}(s' \vert s_{t},a) \nabla_{\theta_{\text{adv}}}\mathbf{V}_{\pi}(s')  \right) \quad\quad  \text{(product rule)} \\
 													&= \sum_a \pi(a \vert s_t)  \left( \sum_{s'} \mathbf{V}_{\pi}(s') \pi_{\text{adv}}(s' \vert s_{t},a)  \frac{\nabla_{\theta_{\text{adv}}} \pi_{\text{adv}}(s' \vert s_{t},a)}{\pi_{\text{adv}}(s' \vert s_{t},a)}  +  \nabla_{\theta_{\text{adv}}}\mathbf{V}_{\pi}(s')  \right) \quad\quad  \\ & \quad \quad\text{(divide and multiply by $\pi_{\text{adv}}$)} \\
 													&= \mathbb{E}_{\pi,\pi_{\text{adv}}}   \left[\mathbf{V}_{\pi}(s_{t+1}) \frac{\nabla_{\theta_{\text{adv}}} \pi_{\text{adv}}(s_{t+1} \vert s_{t},a_t)}{\pi_{\text{adv}}(s_{t+1} \vert s_{t},a_t)}+  \nabla_{\theta_{\text{adv}}}\mathbf{V}_{\pi}(s_{t+1})  \right]\quad\quad  \text{(expectation over $\pi$ and $\pi_{\text{adv}}$ )} \\
 													&= \mathbb{E}_{\pi,\pi_{\text{adv}}} \left[  \mathbf{V}_{\pi}(s_{t+1})\nabla_{\theta_{\text{adv}}} \log\left(\pi_{\text{adv}}(s_{t+1} \vert s_{t},a_t)\right)  +  \nabla_{\theta_{\text{adv}}}\mathbf{V}_{\pi}(s_{t+1})  \right] \,. \quad\quad   \text{(derivative of logarithm)}
\end{align*}
Therefore, expanding this sum across all times $t=0,\dots,T-1$, were $T$ is the horizon of the decision process, the expression for $t=0$ is given by
\begin{align*}
 \nabla_{\theta_{\text{adv}}} \mathbf{V}_{\pi}(s_0) &= \sum_{k=0}^{T-1} \mathbb{E}_{\pi,\pi_{\text{adv}}} \left[ \mathbf{V}_{\pi}(s_{k+1}) \nabla_{\theta_{\text{adv}}} \log\left(\pi_{\text{adv}}(s_{t+1} \vert s_{t},a_t)\right)  \right] \,.
\end{align*}
\qed

\newpage
\section*{Appendix C: Experiment details}
\subsection*{Inventory Management}
For each item, the purchasing cost is $2.49$, the selling price is $3.99$, and the holding cost is $0.03$. The reward $r(s,a)$ is the expected revenue minus the ordering costs and the holding costs. The demand distribution is Gaussian with mean $\mu$ and standard deviation $\sigma$. Each episode consists of $T=100$ steps and the discount is set to $\gamma=0.99$. The constraint-cost is $c(s,a) = \max( 0, a - L(s)$, where the purchasing limit is set to $L(s) = \mu + \sigma$ for $ s \leq 2$ and $L(s) = \mu$ for $ s > 2$. The constraint-cost budget is set to $d=6.0 \approx \sum_{t=0}^{T-1} \gamma^{t} 0.1$ which allows the action to exceed the purchasing limit on average roughly one item every 10 time steps. The constraint-cost function is not adjusted for tests; that is, the original $\mu$ and $\sigma$ are used in its computation rather than the perturbed parameters.

\subsection*{Safe Navigation}
The objective is to move from start, $s_0=(0,0)$ to goal, $(4,4)$, as quickly as possible while 
avoiding areas that incur constraint-costs. The agent observes its $(x,y)$-coordinate and outputs an action going one step left, one step right, one step up, or one step down. The episode is terminated if either the agent arrives at the goal square or if more than $T$ time steps have passed. 
Instead of using the full state space as next states in the uncertainty sets, the probability vectors $\mathcal{P}_{s,a}$ consider for the next 
state only the 5 states in the Von Neumann neighbourhood $\mathcal{N}(s)$ with Manhattan distance of at most 1 from $s$; this requires setting $\alpha(s,a) = \sqrt{\frac{2}{n(s,a)} \ln\left(\frac{2^{S'} S A}{\delta}\right)}$, replacing $S$ by $S'=5$ in the set of outcomes. 

\section*{Appendix D: Training hyperparameters}
\label{appendixC}
\subsection{Hyperparameters}
Hyperparameters are set according to Table~\ref{tab: params}. The discount is common at 0.99 and the architecture was chosen such that it is large enough for both domains. The entropy regularisation is higher than usual training procedures because of the Lagrangian yielding larger numbers in the objective. Learning rates were tuned in $\{0.10,0.01,0.001\}$ for policy parameters ($\theta$ and $\theta_{\text{adv}}$) and in $\{0.01,0.001,0.0001\}$ for Lagrangian multipliers ($\lambda$ and $\lambda_{\text{adv}}$); the setting shown in the table is the best setting for  Inventory Management and Safe Navigation domains and this loosely corresponds to the two time scale stochastic approximation criteria \cite{Borkar2022}. The critic was fixed to 0.001 for both domains as this is a reliable setting for the Adam optimiser. For Inventory Management, it is possible to satisfy the constraint from the initial stages of learning so the initial Lagrangian multiplier $\lambda$ is set to 50. For Safe Navigation domains, the initial $\lambda$ is set to 1 since it is not immediately possible to satisfy the constraints without learning viable paths to goal. To encourage stochasticity in case of limited samples, each state-action pair $(s,a)$ is initialised with a pseudo-count $n(s,a) \gets 1$, representing the uniform distribution as a weak prior belief. The error probability is $\delta=0.10$ for a $90\%$ confidence interval.

\begin{table}
\centering
\caption{Parameter settings of the experiments}\label{tab: params}
\begin{tabular}{l p{6cm}}
\toprule \\ 
\textbf{Parameter} & \textbf{Setting} \\
\midrule
Discount & 0.99 \\
Entropy regularisation for $\pi$ & 5.0 \\
Architecture for $\pi$ and $\pi_{\text{adv}}$ &  100 hidden RELU units, \newline softmax output \\
Learning rates for $\theta$,$\lambda$,$\theta_{\text{adv}}$, and $\lambda_{\text{adv}}$  & 0.001, 0.0001, 0.001, and 0.0001, \newline
 multiplier $\frac{1}{1 + n//500}$ for episode $n$\\
Initialisation of $\lambda$ and $\lambda_{adv}$ & both 50 for Inventory Management, \newline both 1 for Safe Navigation 1 \& 2 \\
Critic & learning rate 0.001, \newline 100 hidden RELU units,\newline linear output,\newline Adam optimisation of MSE,\newline batch is episode  \\ 
Uncertainty set & Hoeffding-based L1, 1 pseudocount, 90\% confidence interval \\
\bottomrule
\end{tabular}
\end{table}
\newpage
\section*{Appendix E: Training and model estimation}
\begin{figure}[htbp!]
\centering
\includegraphics[width=0.65\textwidth]{figures_IM/legend.pdf} \\
\subfloat[Training value]{
\includegraphics[width=0.40\textwidth]{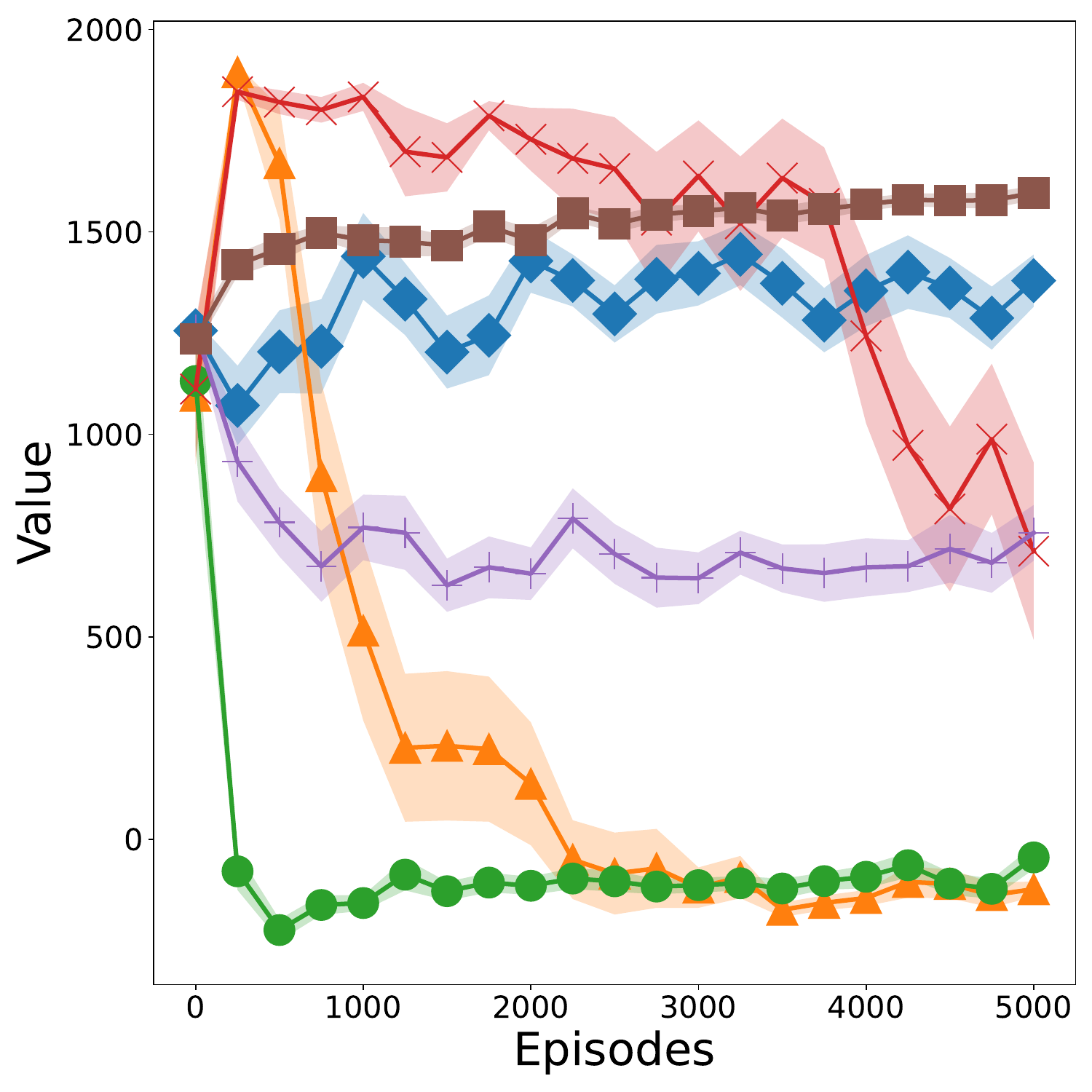}
}
\subfloat[Training overshoot]{\includegraphics[width=0.40\textwidth]{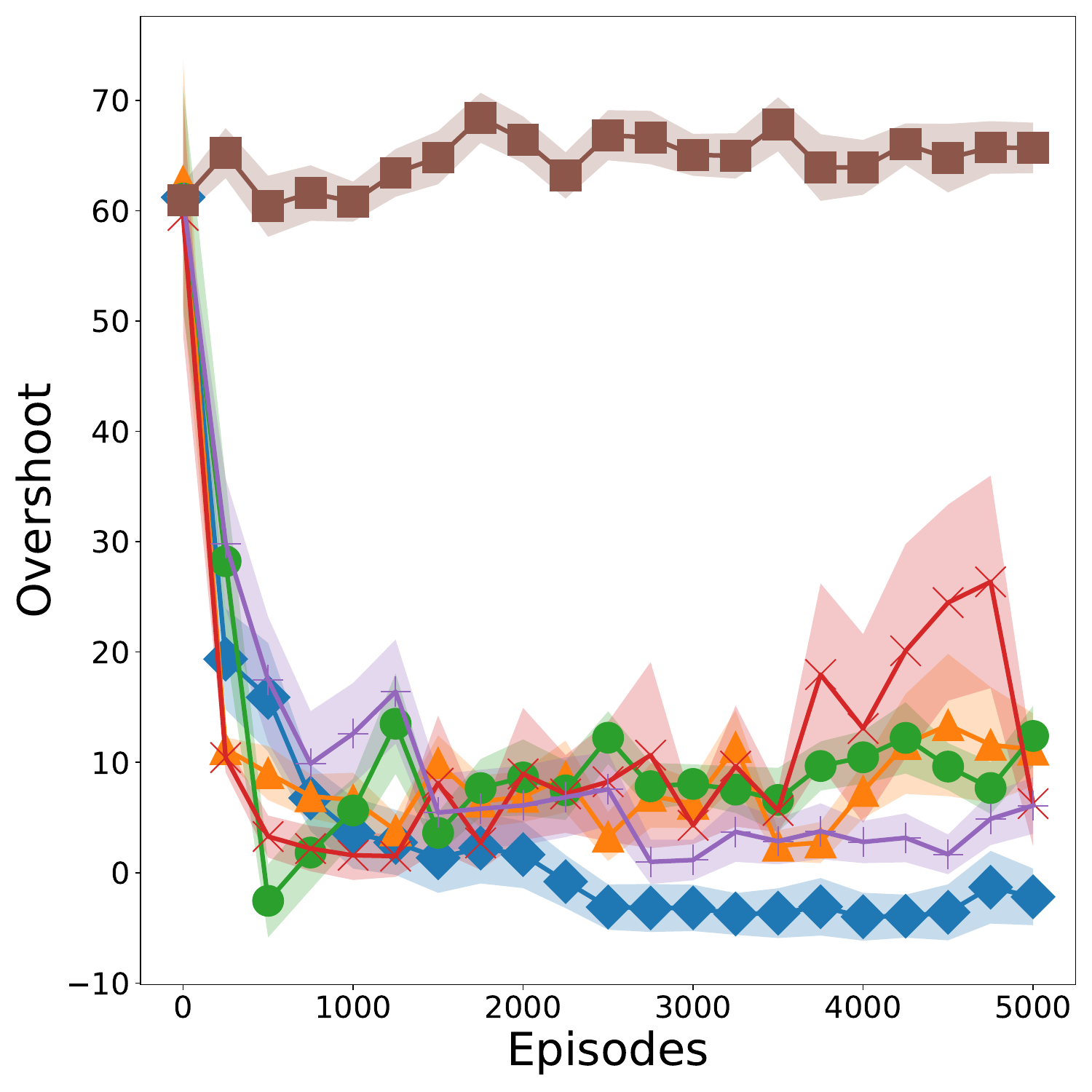}}
\caption{Training performance metrics of the algorithms over 5,000 episodes on Inventory Management. Note that the training performance corresponds to the performance on the simulated transition dynamics, which is defined differently for the different algorithms.} \label{fig: IM development}
\end{figure}

In Inventory Management, the model estimation phase is based on 100 episodes with $\mu = S/4$ and $\sigma = S/6$, yielding uncertainty sets with budget $\alpha$ ranging in 
$[0.3,0.9]$ across the state-action space. The widely varying values and overshoots during training (see Figure~\ref{fig: IM development}) reflect in part a different training environment. Fig.~\ref{fig: IM development} shows the performance in the policy training phase.

\begin{figure}[htbp!]
\centering
\includegraphics[width=0.65\textwidth]{figures/legend.pdf} \\
\subfloat[Training value]{
\includegraphics[width=0.40\textwidth]{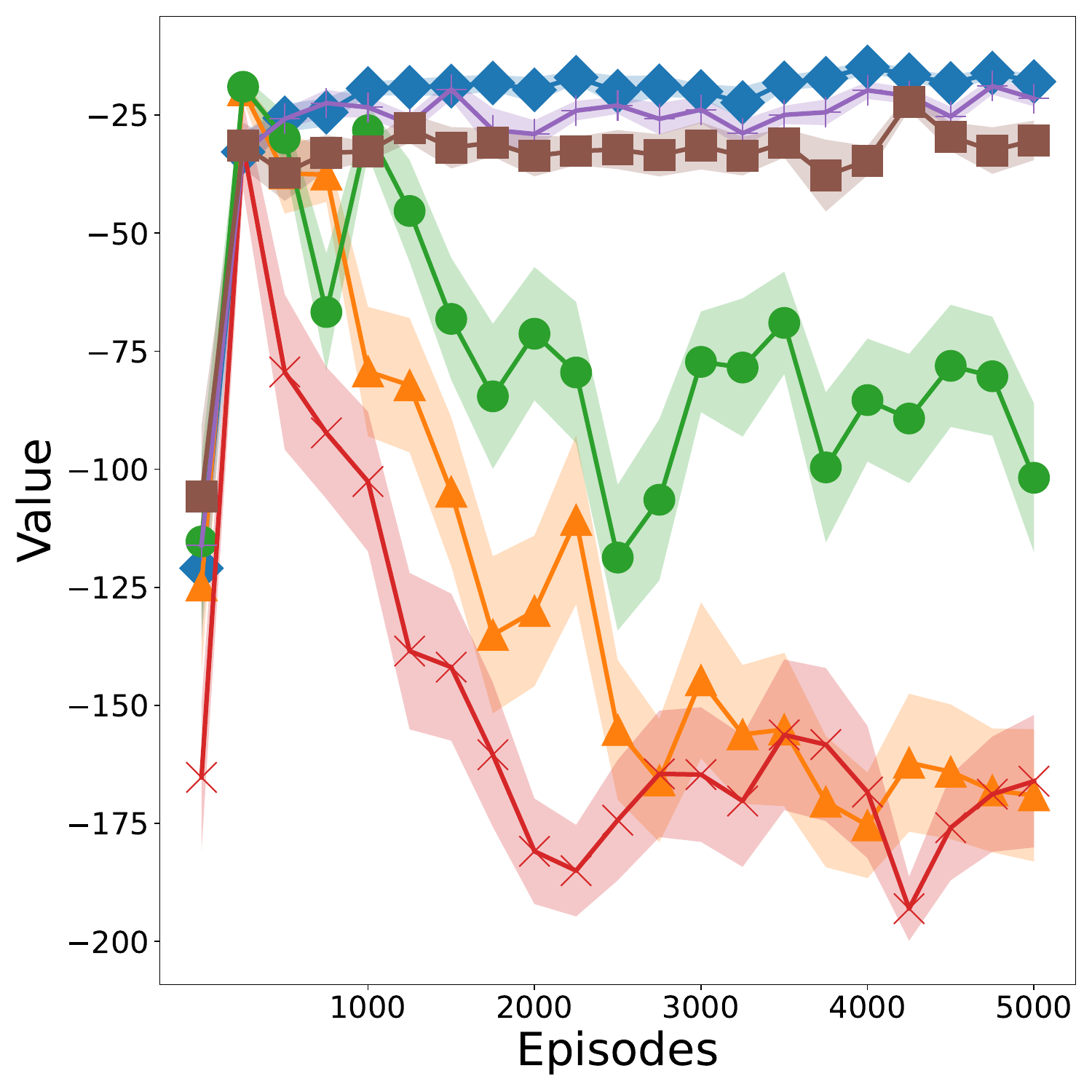}
}
\subfloat[Training overshoot]{\includegraphics[width=0.40\textwidth]{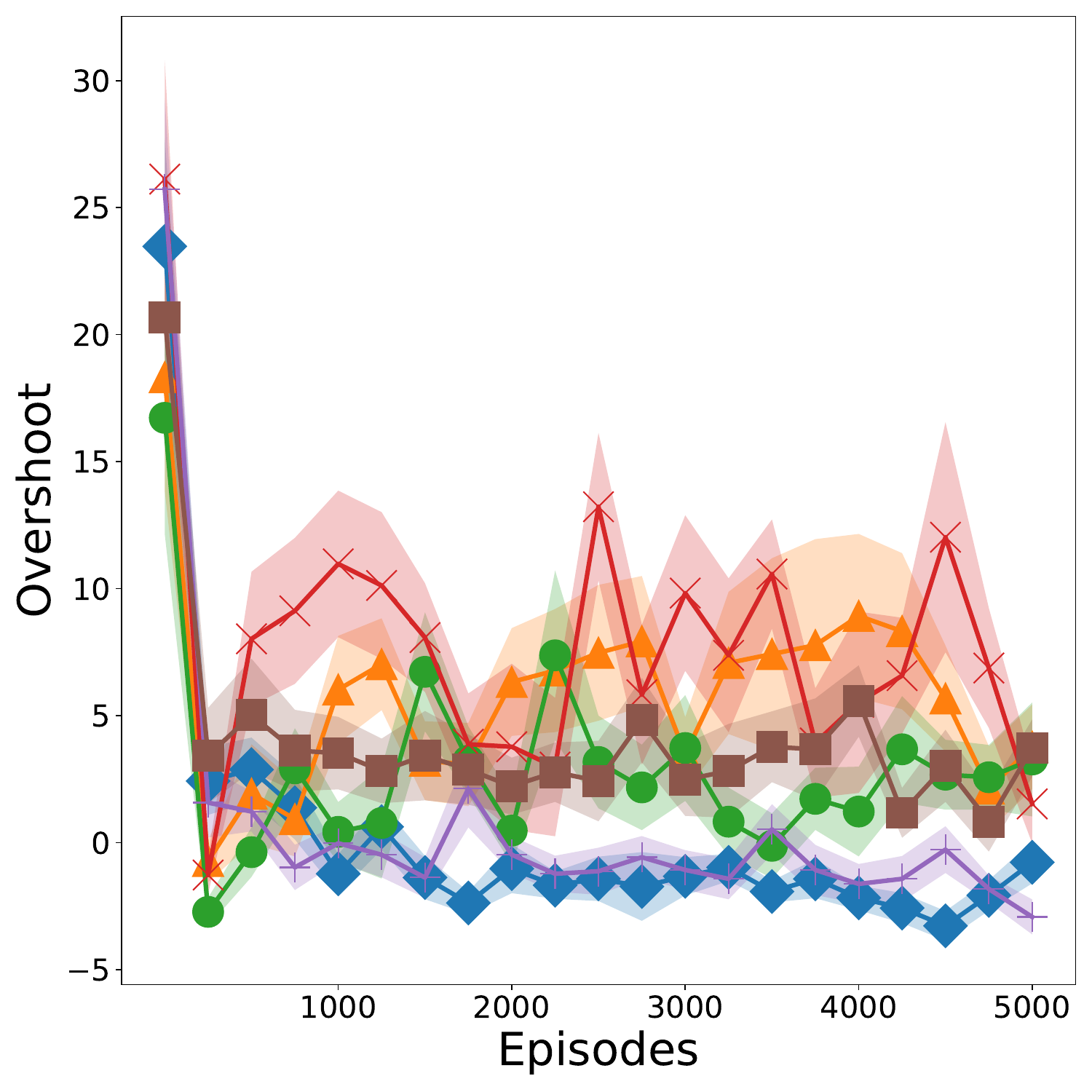}}
\caption{Training performance metrics of the algorithms over 5,000 episodes on Safe Navigation 1. Note that the training performance corresponds to the performance on the simulated transition dynamics, which is defined differently for the different algorithms.} \label{fig: SafeNavigation1 development}
\end{figure}

In Safe Navigation 1, the model estimation phase is based on 100 episodes with $P_{\text{success}}=0.80$, which results in the uncertainty budget $\alpha$ ranging in $[0.25,0.7]$ across the state-action space. Fig.~\ref{fig: SafeNavigation1 development} shows the performance in the policy training phase.

\begin{figure}[htbp!]
\centering
\includegraphics[width=0.65\textwidth]{figures/legend.pdf} \\
\subfloat[Training value]{
\includegraphics[width=0.40\textwidth]{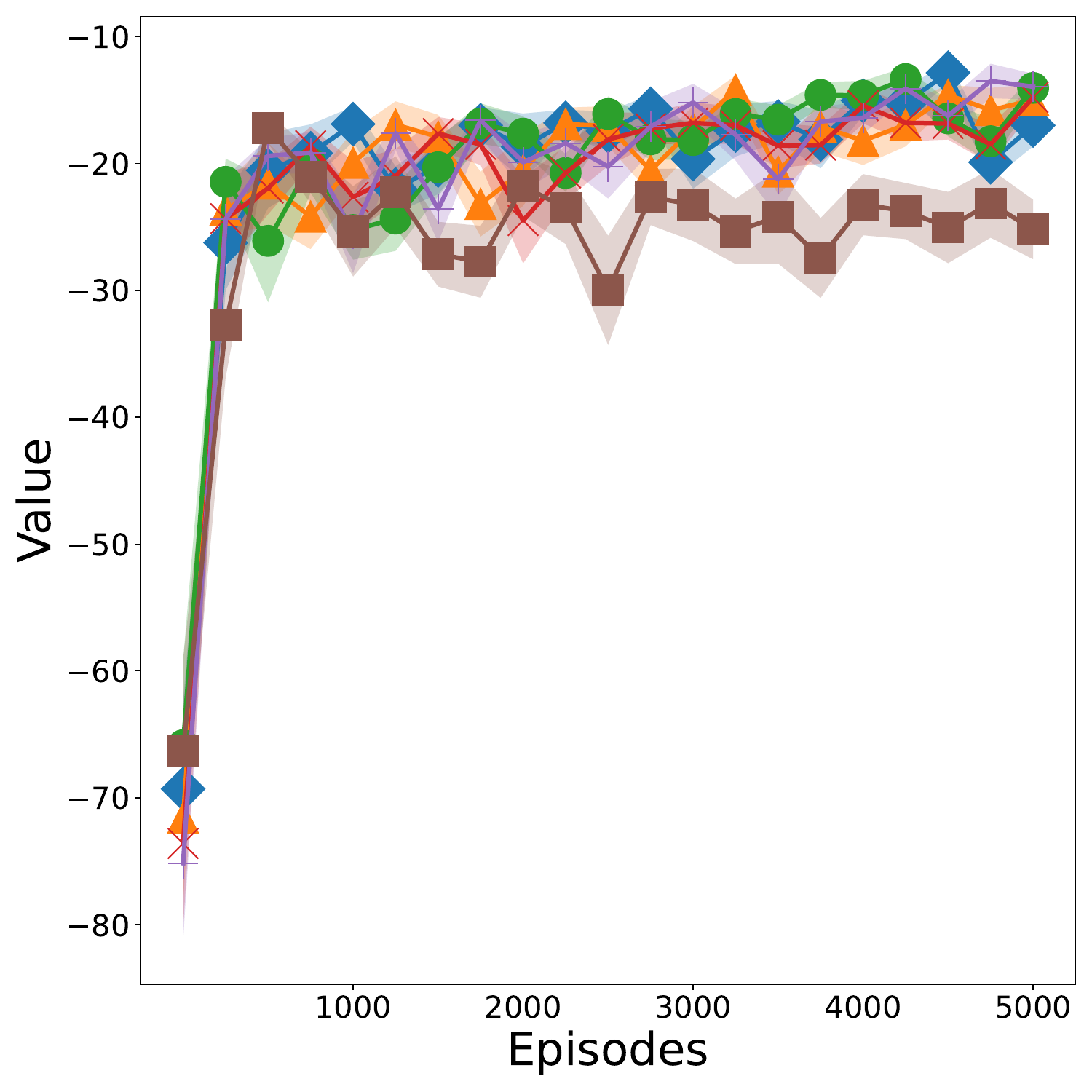}
}
\subfloat[Training overshoot]{\includegraphics[width=0.40\textwidth]{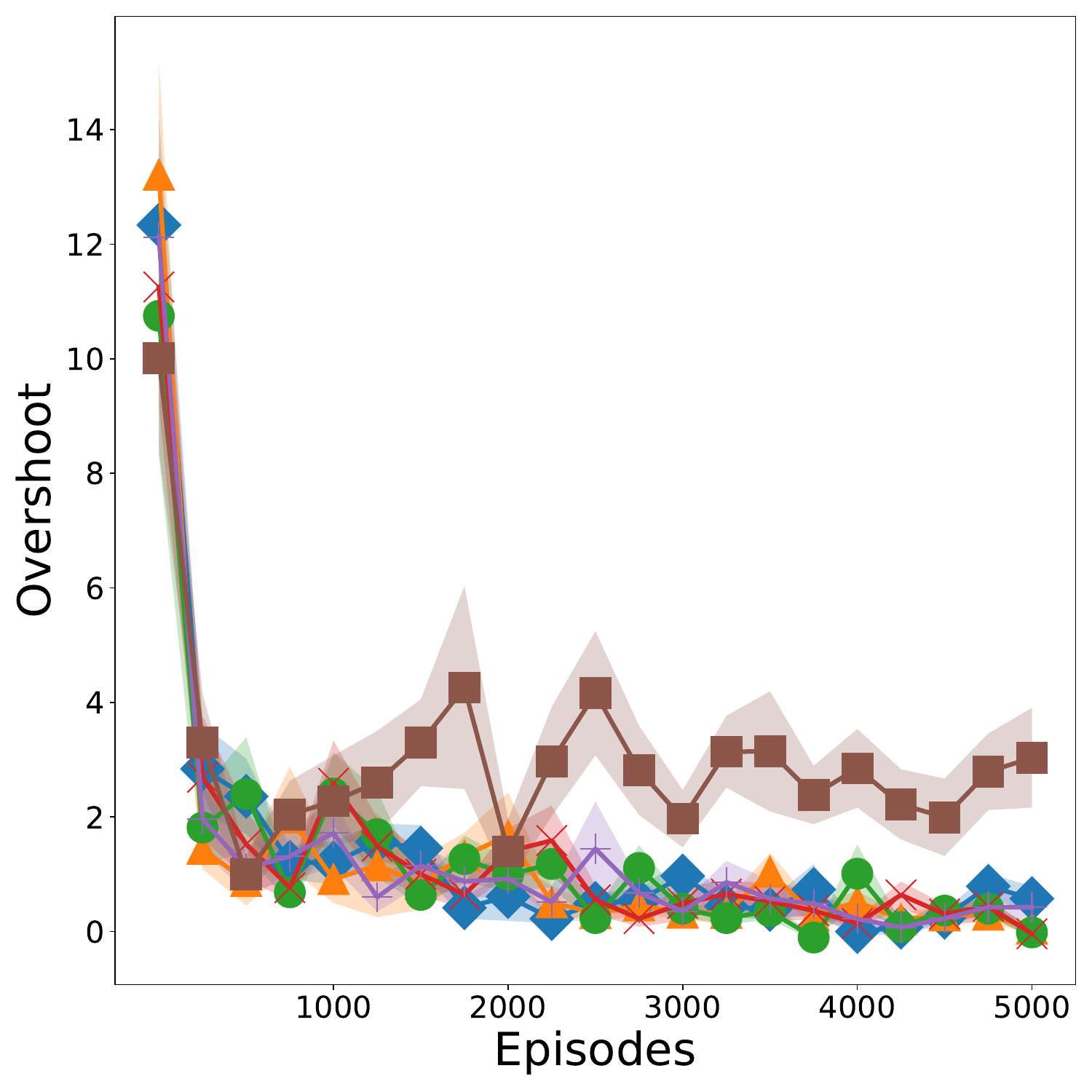}}
\caption{Training performance metrics of the algorithms over 5,000 episodes on Safe Navigation 2. Note that the training performance corresponds to the performance on the simulated transition dynamics, which is defined differently for the different algorithms.}\label{fig: SafeNavigation2 development}
\end{figure}

In Safe Navigation 2, the model estimation phase is based on 10,000 episodes with $P_{\text{success}}=1.0$. The resulting uncertainty set has a smaller uncertainty budget, with  $\alpha$ ranging in $[0.03,0.085]$ across the state-action space. Fig.~\ref{fig: SafeNavigation2 development} shows the performance in the policy training phase.

\end{document}